\documentclass{article}
\PassOptionsToPackage{svgnames,dvipsnames,table}{xcolor}
\usepackage[OT1]{fontenc}
\usepackage{iclr2026_conference,times}

\usepackage{amsmath,amssymb,amsthm}
\usepackage{booktabs,tabularx}
\usepackage{hyperref}
\usepackage{url}
\expandafter\def\expandafter\UrlBreaks\expandafter{\UrlBreaks\do-}
\usepackage{xcolor}
\usepackage{colortbl}
\usepackage{xspace}
\usepackage{graphicx}
\usepackage[most]{tcolorbox}
\usepackage{enumitem}
\usepackage{multirow}
\usepackage{caption}
\usepackage{titletoc}
\usepackage{tocbibind}
\usepackage{marvosym}
\usepackage{pifont}
\usepackage{algorithm,algpseudocode,needspace,placeins}

\definecolor{darkblue}{rgb}{0, 0, 0.5}
\definecolor{citeblue}{rgb}{0.35, 0.62, 0.88}
\definecolor{custompink}{HTML}{FF69B4}
\hypersetup{colorlinks=true, citecolor=citeblue, urlcolor=custompink, linkcolor=red}

\definecolor{rowgray}{gray}{0.95}
\definecolor{caseaccent}{HTML}{315E7D}
\definecolor{caseframe}{HTML}{C9D1D9}
\definecolor{caseback}{HTML}{F4F7FA}
\definecolor{casetask}{HTML}{466A8F}
\definecolor{caseexperience}{HTML}{377A70}
\definecolor{casetrace}{HTML}{8A682D}
\definecolor{casefeedback}{HTML}{745A88}
\definecolor{casefailure}{HTML}{A04F52}
\definecolor{casestrong}{HTML}{3E7654}

\newtcolorbox{promptbox}[1]{
    enhanced,
    breakable=true,
    colback=caseback,
    colframe=caseframe,
    coltitle=caseaccent,
    fonttitle=\bfseries\small,
    title={#1},
    boxrule=0.45pt,
    arc=2pt,
    borderline west={2.2pt}{0pt}{caseaccent},
    attach boxed title to top left={xshift=9pt,yshift=-2.5mm},
    boxed title style={
        colback=caseback,
        colframe=caseback,
        boxrule=0pt,
        left=3pt,
        right=3pt,
        top=1pt,
        bottom=1pt
    },
    left=8pt,
    right=8pt,
    top=6pt,
    bottom=4pt,
    before skip=5pt,
    after skip=5pt
}

\newtcolorbox{casebox}[1]{
    enhanced,
    breakable=false,
    colback=caseback,
    colframe=caseframe,
    coltitle=caseaccent,
    fonttitle=\bfseries\small,
    title={#1},
    boxrule=0.45pt,
    arc=2pt,
    borderline west={2.2pt}{0pt}{caseaccent},
    attach boxed title to top left={xshift=9pt,yshift=-2.5mm},
    boxed title style={
        colback=caseback,
        colframe=caseback,
        boxrule=0pt,
        left=3pt,
        right=3pt,
        top=1pt,
        bottom=1pt
    },
    left=10pt,
    right=9pt,
    top=11pt,
    bottom=7pt,
    before skip=11pt,
    after skip=13pt
}
\newcommand{\casefield}[2]{\noindent\textcolor{#1}{\textbf{#2.}}\enspace}

\theoremstyle{plain}
\newtheorem{proposition}{Proposition}
\newtheorem*{srepropappendix}{Proposition~\ref{prop:biased-reward-exploration}}
\newtheorem{corollary}{Corollary}
\newtheorem*{srecorappendix}{Corollary~\ref{cor:group-centered-gradient}}

\theoremstyle{remark}

\theoremstyle{plain}
\definecolor{methodrowgreen}{RGB}{232,245,233}
\definecolor{tblgroupbg}{RGB}{234,240,247}
\definecolor{tblgreen}{HTML}{2E7D32}
\definecolor{tblred}{HTML}{C62828}
\definecolor{tblamber}{HTML}{B26A00}
\newcommand{\cmark}{\textcolor{tblgreen}{\ding{51}}}
\newcommand{\xmark}{\textcolor{tblred}{\ding{55}}}
\newcommand{\pmark}{\textcolor{tblamber}{$\sim$}}

\newcommand{\btheta}{\boldsymbol{\theta}}

\newcommand{\github}{\raisebox{-1.5pt}{\includegraphics[height=1.05em]{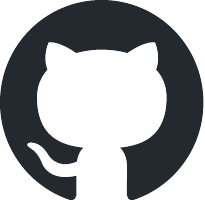}}}
\newcommand{\huggingface}{\raisebox{-1.5pt}{\includegraphics[height=1.05em]{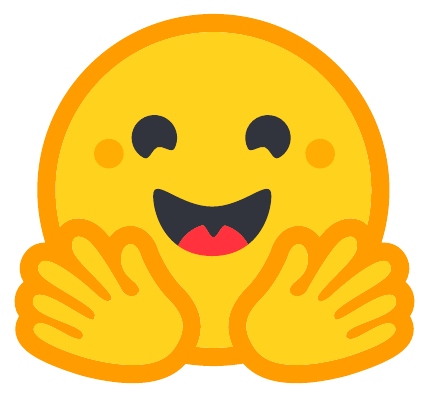}}}

\makeatletter
\def\@fnsymbol#1{%
  \ifcase#1 \or
    *%
  \or
    \textrm{\Letter}%
  \or
    \ddagger
  \or
    \S
  \or
    \P
  \or
    \|
  \or
    **%
  \or
    \dagger\dagger
  \or
    \ddagger\ddagger
  \else
    \@ctrerr
  \fi
}
\makeatother

\title{Agent Learning from Single-Feedback Deployment~Experience}

\author{\textbf{Rong-Xi Tan}$^{1,2}$\thanks{~Equal contribution.}\kern0.45em, \enspace
\textbf{Yizhuo Li}$^{3,2}$\footnotemark[1]\kern0.45em, \enspace
\textbf{Wenjie Wang}$^{4}$\footnotemark[1]\kern0.45em, \enspace
\textbf{Xiaoye Qu}$^{2}$\\
\textbf{Qiwei Chen}$^{5}$, \enspace
\textbf{Tong Zhu}$^{2}$, \enspace
\textbf{Yun Luo}$^{2}$, \enspace
\textbf{Zhi Wang}$^{1}$\\
\textbf{Chao Qian}$^{1}$\thanks{~Corresponding authors.}\kern0.65em, \enspace
\textbf{Yu Cheng}$^{6}$\footnotemark[2]\kern0.65em, \enspace
\textbf{Yang Yang}$^{3}$\footnotemark[2]\kern0.65em, \enspace
\textbf{Yafu Li}$^{7,2}$\footnotemark[2]\kern0.65em\\[4pt]
\normalfont $^{1}$Nanjing University \quad $^{2}$Shanghai Artificial Intelligence Laboratory\\
\normalfont $^{3}$Shanghai Jiao Tong University \quad $^{4}$Beijing Jiaotong University\\
\normalfont $^{5}$Harbin Institute of Technology \quad $^{6}$Nanyang Technological University\\
\normalfont $^{7}$The Chinese University of Hong Kong
}

\iclrfinalcopy
\begin{document}

\maketitle
\vspace{-0.65em}
\begin{abstract}
The success of LLM agents has attracted growing interest in deploying small-size, self-hosted models to serve as user assistants for everyday work.
After deployment, users would like these assistants to improve as they work.
In practice, however, tasks arrive sequentially, and users cannot be expected to repeatedly evaluate different attempts at the same task.
A common approach is to adapt memories or skills in the agent harness while keeping model parameters fixed.
Parameter-based learning provides another promising direction, yet how to effectively train models from such single-feedback deployment experience remains under-explored.
In this paper, we study this problem and introduce DEAL, an executable testbed with related task streams in general workspace assistance and AI research.
Using DEAL, we find that single-rollout reinforcement learning suffers from an exploration bottleneck, where one evaluated trajectory may leave better behaviors unexplored.
Based on this observation, we propose Self-Rewarded Exploration (SRE), which expands additional trajectories from isolated task snapshots and uses the live feedback as an anchor to assign self-rewards.
We further provide a theoretical analysis of when the benefit of rollout expansion can outweigh self-reward errors.
Experiments on Gemma4-E4B-IT and Qwen3.5-9B demonstrate the effectiveness of SRE in improving deployment-time performance, generalizing to unseen tasks, and preserving general capabilities.
\vspace{-0.5em}
\end{abstract}

\begin{center}
\vspace{-0.3em}
~\github~\href{https://github.com/trxcc/SRE}{\textcolor{custompink}{Code}}
\quad \quad \quad
~\huggingface~\href{https://huggingface.co/collections/trxcc2002/sre-deal}{\textcolor{custompink}{Datasets}}
\quad \quad \quad
~\huggingface~\href{https://huggingface.co/collections/trxcc2002/sre-deal}{\textcolor{custompink}{Models}}
\vspace{0.2em}
\end{center}

\section{Introduction}
\label{sec:introduction}

The success of Large Language Model (LLM) agents has attracted growing interest in deploying small, self-hosted models as user assistants for everyday work~\citep{belcak2025smalllanguagemodelsfuture,minicpm4}.
Through an agent harness, these assistants interact with files and tools to analyze data, modify code, and prepare reports~\citep{openclaw2026}.
After deployment, users may also expect them to improve as they work~\citep{ren2026selfimprovementsmodernagenticsystems}.
Task outcomes and user feedback provide useful experience about recurring mistakes or user requirements~\citep{wang2026openclaw, zhang2026pibenchevaluatingproactivepersonal}.

Learning after deployment differs from the gym-based agentic training before deployment. 
In a training gym, agents can revisit tasks and receive feedback on new attempts. 
During deployment, tasks arrive along the user's workflow, and users may move on after evaluating one delivered result. 
The agent must therefore use the limited experience and feedback from completed tasks to improve its performance on subsequent tasks~\citep{guo2026cascade}.
We study this process of \textit{deployment-time self-evolution}, in which agents adapt their model parameters or persistent harness state through continued use.
Specifically, there exists a task stream where the agents can only complete one live execution and receive one external outcome feedback per task. 
Existing work explores different ways to use this experience.
One direction updates memories or reusable skills while keeping the model fixed~\citep{chhikara2025mem0buildingproductionreadyai,agrawal2026gepareflectivepromptevolution,alzubi2026evoskill,ma2026skillclawletskillsevolve}, while another updates model parameters through live interactions~\citep{wang2026openclaw,xia2026metaclawjusttalk}.
Both approaches aim to turn deployment experience into transferrable signals on subsequent tasks.

Evaluating deployment-time self-evolution requires both realistic assistant tasks and opportunities to reuse experience. 
Agents should complete work through a harness, while different tasks should share materials, requirements, or working procedures~\citep{asawa2026continuallearningbenchevaluating}. 
Existing benchmarks provide useful task resources~\citep{bai2026clawgymscalableframeworkbuilding, ji2026clawarenabenchmarkingaiagents, guo2026cascade}, but further task construction is needed to support long streams of related tasks under a common deployment-time protocol. 
We therefore introduce DEAL (\underline{D}eployment-time \underline{E}valuation of \underline{A}gent \underline{L}earning), an executable testbed covering \textit{general workspace assistance} and \textit{AI research}. 
We construct related tasks from workspace files and research materials, and refine them through execution and quality checks. 
In DEAL, shared abilities and work contexts connect the tasks, while rubric-based user simulation provides feedback on the execution artifacts.
DEAL adopts a single-feedback protocol to study learning with limited user supervision, where the agent receives one user feedback for each online task.

Under DEAL, we implement a range of baselines covering both training-free methods and parameter-based learning. We further study parameter-based learning, as recent work shows that parameter updates can help models internalize experience and improve task execution~\citep{xia2026metaclawjusttalk,harrington2026doescontinuallearningrequire}. Under the single-feedback constraint, an intuitive approach is to train the model from live trajectories and their rewards through single-rollout reinforcement learning (RL)~\citep{hou2026sao,hu2026flashreinforce}. 
However, the limited feedback may reveal problems in the delivered result without providing a better trajectory to learn from. 
Whether the observed result is good or poor, alternative behaviors that may provide advantage signals remain unexplored. 
We refer to this limitation as an \textit{exploration bottleneck}.

In this paper, we point out that one user feedback need not limit learning to one trajectory, and the agent can asynchronously get more rollouts in free time.
Based on this observation, we propose Self-Rewarded Exploration (SRE).
After live execution, SRE generates additional trajectories in isolated copies of the saved environment snapshots.
These candidates receive no new user evaluations.
Instead, SRE uses the live trajectory and its feedback as an anchor to construct a shared rubric and assign self-rewards.
The internal trajectories and their self-rewards are then used for GRPO-style updates~\citep{shao2024deepseekmath}. Each update group uses only self-rewards, as mixing them with external scores may distort the advantages.
The additional rollouts provide learning experience for future tasks without replacing the current delivery.
We further theoretically analyze when additional rollouts yield better candidates despite biased self-rewards.

We conduct experiments with Gemma4-E4B-IT~\citep{gemmateam2026gemma4} and Qwen3.5-9B~\citep{qwen3.5}, comparing SRE with CASCADE~\citep{guo2026cascade}, EvoSkill~\citep{alzubi2026evoskill}, and REINFORCE++~\citep{hu2025reinforcepp} on DEAL, as well as compute-matched RL and feedback-conditioned distillation~\citep{hubotter2026reinforcementlearningselfdistillation}.
SRE obtains the largest improvement of deployment-time performance and unseen task-solving abilities in both workspace and research domains, while preserving general capabilities.
Further diagnostics and ablations show that additional rollouts provide better candidates and that the feedback anchor reduces selection errors.
We also find that mixing external scores with self-rewards can distort group-relative advantages, supporting our use of internal-only reward groups.

Our contributions are summarized as follows:
\begin{itemize}
    \item We introduce DEAL, an executable testbed for deployment-time self-evolution on related workspace and AI research tasks, combining harness-based execution, shared task structure, and a single-feedback protocol across training-free and parameter-based methods.

    \item We identify an \textit{exploration bottleneck} in learning from a single live trajectory and propose SRE, which combines internal rollout expansion with feedback-conditioned self-rewards to support parameter learning without additional user evaluations.

    \item We characterize the trade-off between exploration gain and selection error, bound the effect of self-reward errors on group-centered policy gradients, and validate SRE across models, rollout budgets, and imperfect feedback settings.
\end{itemize}

\section{Agent Deployment Scenario and Learning Protocol}
\label{sec:scenario}

\begin{figure}[t]
    \centering
    \includegraphics[width=\linewidth]{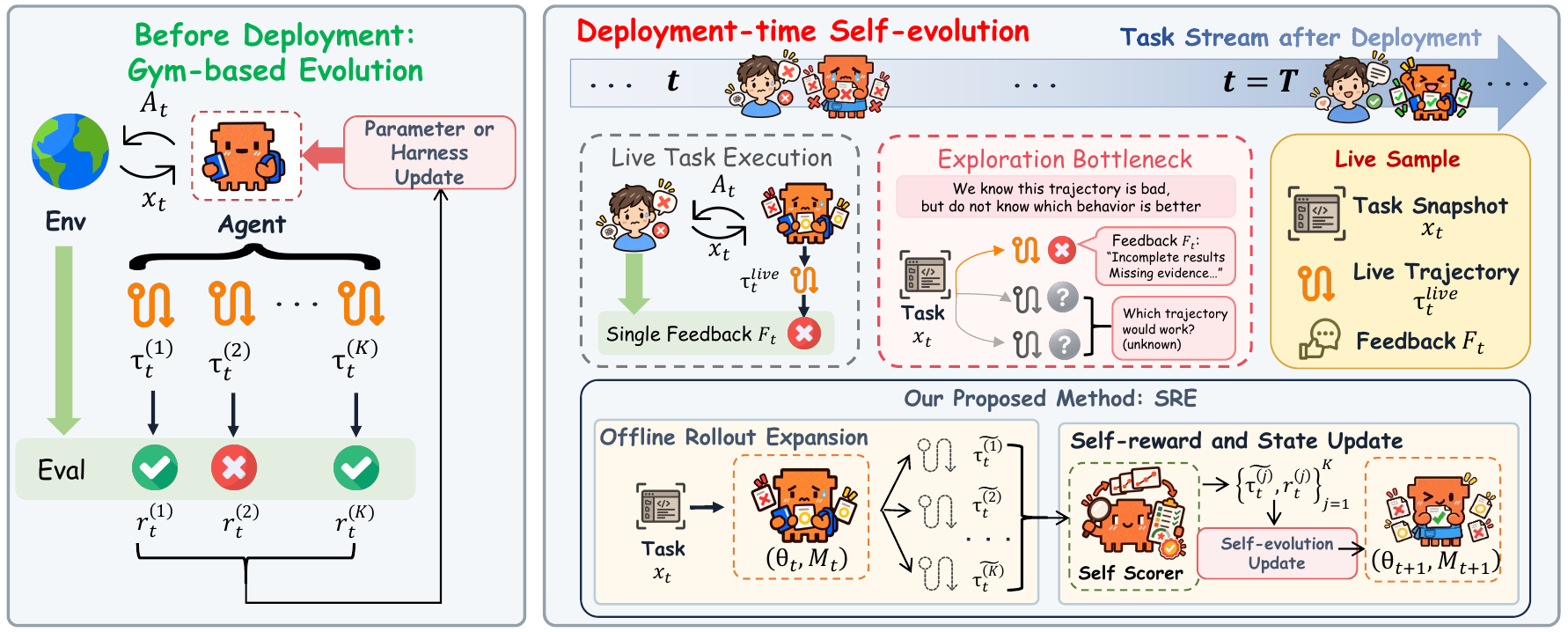}
    \caption{
        Deployment-time self-evolution from a stream of user tasks.
        For each task, the agent completes the single live execution and receives user feedback.
        It then uses the accumulated experience to update its states.
        The objective is to maximize the user's satisfaction continually.
    }
    \label{fig:deployment-protocol}
\end{figure}

In this section, we introduce the background of this paper, i.e., deployed agents that assist users with everyday work.
These agents interact with files and tools through a harness, and learn from completed tasks to improve their future performance, which we term \textit{deployment-time self-evolution}.
Figure~\ref{fig:deployment-protocol} shows the difference between deployment-time self-evolution and  traditional gym-based evolution, where tasks can be repeatedly executed and evaluated.
During deployment, tasks instead follow the user's workflow, and users may move on after providing feedback, prohibiting further live attempts~\citep{guo2026cascade}.
Below, we introduce the deployment-time self-evolution protocol and formalize its objective.

\subsection{Deployment-time Self-evolution}
\label{sec:interaction-protocol}

Let $\mathcal{X}=(x_1, x_2, \cdots, x_T)$ be the task stream where $x_t$ is the $t$-th arriving task.
Denote $s_t$ as the initial environment state, and $(\btheta_{t-1},M_{t-1})$ as the agent's state before that task, consisting of model weights and persistent harness state (e.g., memories or skills).
The initial agent state is $(\btheta_0,M_0)$.
For each task, the agent completes one live trajectory and then receives one outcome feedback:
\begin{align*}
    \tau_t
    &\sim \pi_{\btheta_{t-1}}
    (\cdot \mid x_t,s_t,M_{t-1}), \quad
    F_t
    \leftarrow
    \operatorname{ExternalFeedback}(x_t,\tau_t).
\end{align*}
Here, $\tau_t$ contains the complete multi-turn execution, including model actions, tool observations, and the delivered result.
The execution history available after task $x_t$ is
$    \mathcal H_t
    =
\{(x_k,s_k,\tau_k,F_k)\}_{k=1}^{t}$.
After receiving \(F_t\), the agent can adapt its state using \(\mathcal H_t\):
\begin{equation*}
    (\btheta_t,M_t)
    =
    \mathcal U(\btheta_{t-1},M_{t-1},\mathcal H_t),
    \label{eq:update}
\end{equation*}
where $\mathcal U$ denotes the self-evolution procedure.
Self-evolution $\mathcal U$ can change an agent's persistent context and harness, i.e., updating $M_t$ while keeping the model parameters fixed~\citep{karten2026continualharnessonlineadaptation,zhang2026agenticcontextengineeringevolving}, its parameters, i.e., updating $\btheta_t$~\citep{wang2026openclaw, lei2026deltamemefficientonlinememory}, or both its harness and weights~\citep{hebbar2026siaselfimprovingai,xia2026metaclawjusttalk}.

\subsection{Deployment-time Learning Objective}
\label{sec:objective}

To measure the value of deployment-time self-evolution, we compare the learning agent with a frozen base agent over the same task stream $\mathcal X$.
The comparison quantifies the benefit of learning throughout deployment, rather than only the performance of the final agent.

Let $u_t(\tau)$ denote the utility of the trajectory $\tau$ in the task $x_t$.
We first collect the base trajectory \(\tau_t^{\mathrm{base}}\sim\pi_{\btheta_0}(\cdot\mid x_t,s_t,M_0)\) using the fixed initial agent state.
Motivated by~\citet{guo2026cascade, asawa2026continuallearningbenchevaluating}, the objective is to design a self-evolution method $\mathcal U$ that maximizes the expected average utility gain over the frozen base agent:
\begin{equation}
    J_T(\mathcal U)
    = \frac1T
    \sum_{t=1}^{T}
    \mathbb{E}_{\tau_t, \tau_t^{\mathrm{base}}}
    \left[
        u_t(\tau_t)
        -
        u_t(\tau_t^{\mathrm{base}})
    \right].
    \label{eq:objective}
\end{equation}
A positive $J_T(\mathcal U)>0$ indicates an average gain over the frozen base during deployment, and a higher value of $J_T(\mathcal U)$ represents a better gain.
To rigorously evaluate the self-evolving agents after deployment, in the next section, we introduce a testbed within two everyday scenarios, specifying the task environments and evaluation protocol.

\section{DEAL: A Deployment-time Self-evolution Testbed}
\label{sec:testbed}
In this section, we introduce DEAL (\underline{D}eployment-time \underline{E}valuation of \underline{A}gent \underline{L}earning), a testbed for studying deployment-time self-evolution on two domains, \textit{workspace assistance} and \textit{AI research}.

\subsection{Motivation}
\begin{table}[!b]
    \centering
    \vspace{-0.4em}
    \caption{
        Comparison of DEAL with representative agent benchmarks (\cmark: supported, \pmark: partial, \xmark: unsupported).
        DEAL totals 4,200 tasks across both domains, including 1,800 online and 200 test tasks in DEAL-RS, and 2,000 online and 200 test tasks in DEAL-WS.
    }
    \label{tab:bench-comparison}
    \scriptsize
    \setlength{\tabcolsep}{2.6pt}
    \renewcommand{\arraystretch}{1.08}
    \vspace{0.25em}
    \resizebox{\textwidth}{!}{%
    \begin{tabular}{llccccccr}
        \toprule
        \textbf{Benchmark}
        & \textbf{Domain}
        & \shortstack{\textbf{Harness-}\\\textbf{Native}}
        & \shortstack{\textbf{File}\\\textbf{Workspace}}
        & \shortstack{\textbf{Artifact}\\\textbf{Deliver.}}
        & \shortstack{\textbf{Shared}\\\textbf{Struct.}}
        & \shortstack{\textbf{Deploy}\\\textbf{Stream}}
        & \shortstack{\textbf{Single}\\\textbf{Feedback}}
        & \textbf{\#Tasks} \\
        \midrule
        \rowcolor{tblgroupbg}\multicolumn{9}{l}{\textbf{\textit{(i) Flagship workspace, assistant \& coding agent benchmarks}}} \\
        GAIA~\citep{mialon2024gaia} & General AI assistant & \pmark & \pmark & \xmark & \xmark & \xmark & \xmark & 466 \\
        SWE-bench~\citep{jimenez2024swebench} & Software engineering & \pmark & \cmark & \pmark & \pmark & \xmark & \xmark & 2,294 \\
        OSWorld~\citep{xie2024osworld} & Desktop OS \& apps & \cmark & \cmark & \cmark & \xmark & \xmark & \xmark & 369 \\
        $\tau^2$-Bench~\citep{barres2026tau2bench} & Dual-control tool use & \xmark & \xmark & \xmark & \pmark & \xmark & \xmark & 278 \\
        Terminal-Bench~\citep{merrill2026terminalbench} & Hard CLI / terminal & \cmark & \cmark & \cmark & \xmark & \xmark & \xmark & 89 \\
        Workspace-Bench~\citep{tang2026workspacebench10benchmarkingai} & Enterprise workspace & \cmark & \cmark & \cmark & \pmark & \xmark & \xmark & 388 \\
        \midrule
        \rowcolor{tblgroupbg}\multicolumn{9}{l}{\textbf{\textit{(ii) Flagship AI research \& ML engineering benchmarks}}} \\
        ScienceAgentBench~\citep{chen2025scienceagentbench} & Scientific data \& code & \cmark & \cmark & \pmark & \xmark & \xmark & \xmark & 102 \\
        MLE-bench~\citep{chan2025mlebench} & Kaggle ML engineering & \cmark & \cmark & \pmark & \xmark & \xmark & \xmark & 75 \\
        PaperBench~\citep{starace2025paperbench} & AI paper replication & \cmark & \cmark & \cmark & \xmark & \xmark & \xmark & 20 \\
        \midrule
        \rowcolor{tblgroupbg}\multicolumn{9}{l}{\textbf{\textit{(iii) Memory, continual \& deployment-time adaptation benchmarks}}} \\
        LongMemEval~\citep{wu2025longmemeval} & Multi-session memory & \xmark & \xmark & \xmark & \cmark & \pmark & \xmark & 500 \\
        CL-Bench~\citep{asawa2026continuallearningbenchevaluating} & 6 stateful domains & \pmark & \pmark & \pmark & \cmark & \cmark & \cmark & 301 \\
        MetaClaw-Bench~\citep{xia2026metaclawjusttalk} & Workday CLI / QA & \cmark & \pmark & \pmark & \cmark & \cmark & \cmark & 934 \\
        \midrule
        \rowcolor{methodrowgreen}
        \textbf{DEAL (Ours)} & \textbf{Workspace + AI Research} & \cmark & \cmark & \cmark & \cmark & \cmark & \cmark & \textbf{4,200} \\
        \bottomrule
    \end{tabular}%
    }
    \vspace{-0.4em}
\end{table}

Evaluating deployment-time self-evolution requires a testbed that reflects how assistants work and learn during use.
Users typically deploy models through a harness for tool use and task execution~\citep{openai2025codex,openclaw2026}.
Their needs also extend beyond question answering to completing tasks in a workspace.
Meanwhile, feedback from completed tasks can only benefit subsequent work.
The task stream therefore needs opportunities to reuse earlier experience~\citep{pmlr-v54-alquier17a,asawa2026continuallearningbenchevaluating}.
These considerations lead to three key requirements for testbeds:
\begin{enumerate}[topsep=0.0pt, partopsep=0pt]
    \item [(1)] The agent should be wrapped in a harness system like Codex~\citep{openai2025codex};
    \item [(2)] The tasks should reflect realistic assistant workflows, requiring agents to interact with files and tools in a workspace to produce artifacts, rather than merely answering questions;
    \item [(3)] The task stream should share structural biases to transfer previous knowledge to later tasks~\citep{guo2026cascade,shu2026agentclrigorousevaluationcontinual,harrington2026doescontinuallearningrequire}. 
\end{enumerate}

Table~\ref{tab:bench-comparison} compares DEAL with representative agent benchmarks across harness-based execution, file workspaces, artifact deliverables, shared task structure, online task streams, and single-feedback supervision.
Existing workspace, coding, and research benchmarks provide realistic file environments, but evaluate tasks as independent episodes rather than an online stream.
Recent benchmarks for memory and continual adaptation~\citep{wu2025longmemeval,asawa2026continuallearningbenchevaluating,guo2026cascade,ji2026clawarenabenchmarkingaiagents,xia2026metaclawjusttalk} evaluate sequential task streams, where Continual Learning Bench (CL-Bench) and MetaClaw also use single per-task feedback.
However, their tasks mainly involve state queries, API interactions, or lightweight checks rather than multi-file workspace deliverables.
We therefore introduce DEAL to combine harness-based execution, file-grounded deliverables, shared task structure, and single-feedback evaluation in one testbed.

\subsection{Task Construction and Overview}
\label{sec:construction}

\begin{figure}[!t]
    \centering
    \includegraphics[trim=9.5pt 0pt 6pt 0pt, clip, width=\linewidth]{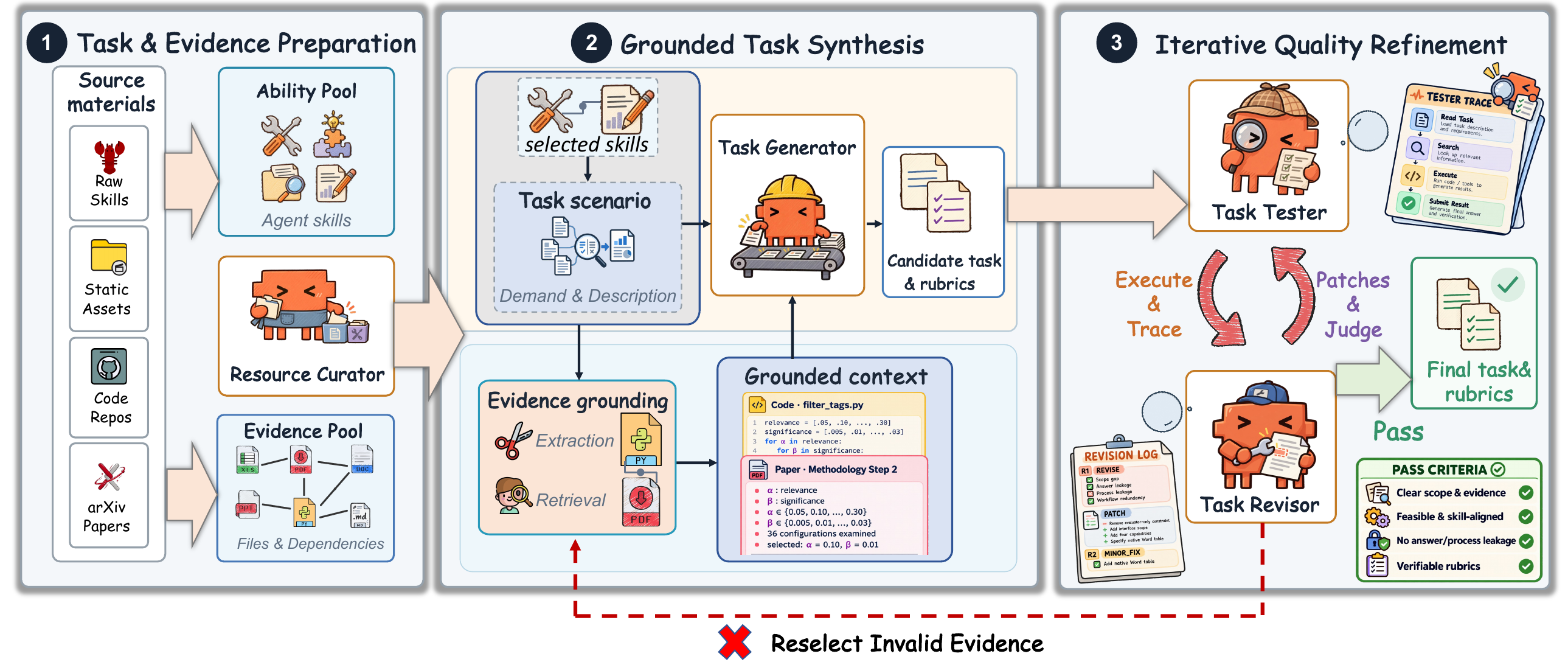}
    \vspace{-0.9em}
    \caption{
        The shared task construction pipeline.
        Source materials are matched with capability requirements and evaluation rubrics to generate tasks with the resources needed for execution and the criteria used to evaluate the work.
    }
    \label{fig:data-construction}
    \vspace{-0.6em}
\end{figure}

Inspired by prior work on agentic data construction~\citep{kimiteam2026kimik3,bai2026clawgymscalableframeworkbuilding,hua2026cliuniverseverifiabletasksynthesis}, we combine grounded task synthesis with execution-based refinement.
Figure~\ref{fig:data-construction} summarizes our three-stage pipeline, with all construction agents operating through a harness.

We begin with \textbf{Task and Evidence Preparation}. 
The resource curator agent organizes source materials into an ability pool of reusable capabilities and an evidence pool of files and dependencies. 
Abilities are shared and recombined across tasks to support cross-task learning.
In \textbf{Grounded Task Synthesis}, the curator turns selected abilities into task scenarios and retrieves supporting evidence. The task generator agent then uses these inputs to create candidate tasks and evaluation rubrics.
Finally, in \textbf{Iterative Quality Refinement}, the task tester executes candidates for rubric-based evaluation.
The task revisor uses execution and judging traces to revise the tasks and rubrics for retesting.
Cases with invalid or insufficient evidence return to the resource curator.     
Tasks are accepted for four criteria: clear scope and evidence, feasibility and skill alignment, no answer or process leakage, and verifiable rubrics. 
We also manually review tasks and rubrics against these criteria and return identified problems for revision. 
Construction details and model configurations are provided in Appendix~\ref{app:task-construction}.

\begin{figure*}[t]
    \centering
    \includegraphics[width=\textwidth]{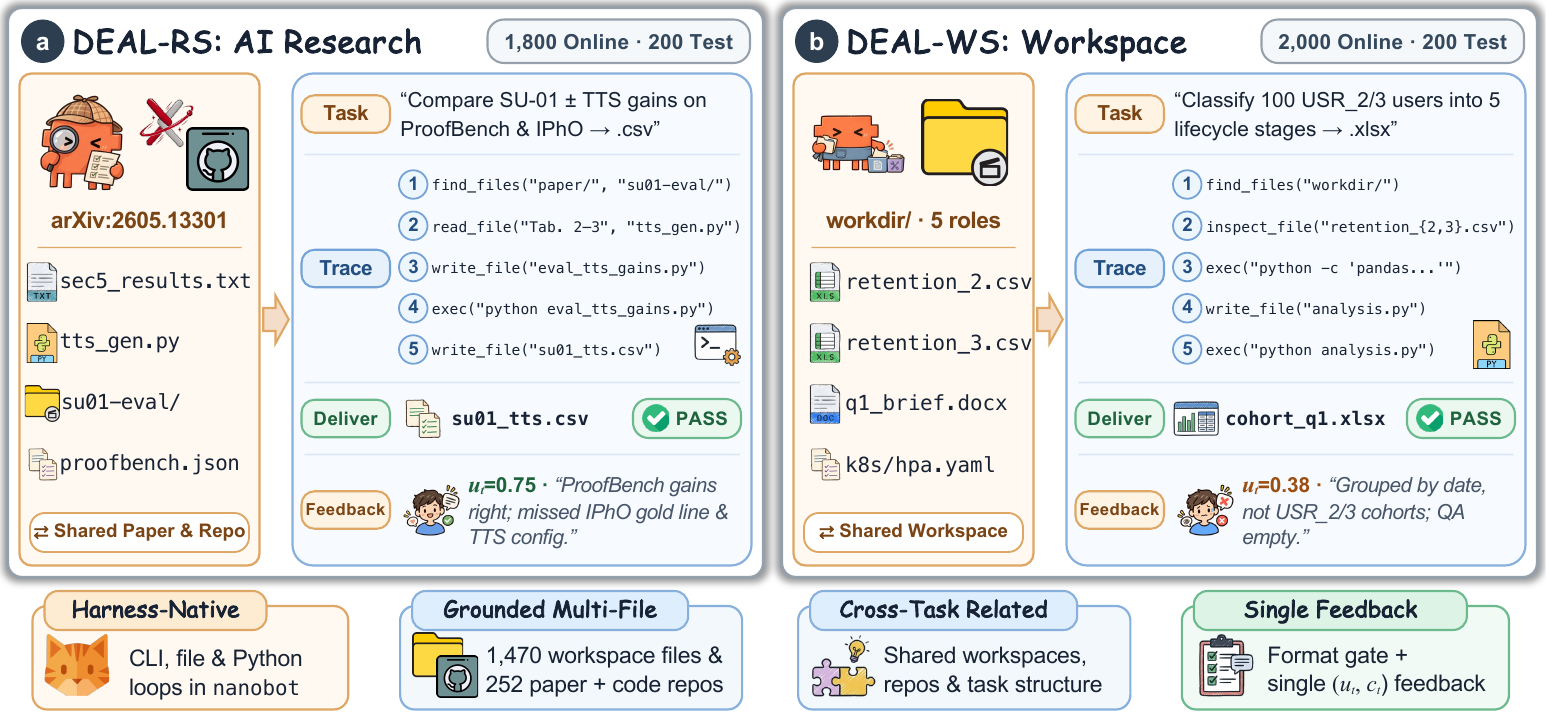}
    \vspace{-0.7em}
    \caption{
        Task demonstrations of DEAL in (a) AI research (DEAL-RS, grounded in SU-01~\citep{li2026su01}) and (b) workspace assistance (DEAL-WS), showing the grounded files, multi-step execution, and single-feedback evaluation alongside four core properties.
    }
    \label{fig:deal-demo}
    \vspace{-0.5em}
\end{figure*}

We instantiate DEAL in two assistant-work scenarios: \textit{AI Research} (DEAL-RS) and \textit{General Workspace Assistance} (DEAL-WS), as illustrated in Figure~\ref{fig:deal-demo}.

\paragraph{DEAL-RS.} 
We screen all papers posted to arXiv's \texttt{cs.AI} category between September 2025 and June 2026 and retain the 252 papers with associated open-source code repositories (18,274 files).
The initial ability pool is built from AI research skills selected from ClawHub and SkillsHub.
We combine these abilities with evidence to construct 1,800 online tasks.
Figure~\ref{fig:deal-demo}(a) shows an example task in DEAL-RS, where the agent reads experimental tables and evaluation code, runs a Python script to evaluate test-time scaling gains, and writes the structured output (\texttt{su01\_tts.csv}).
    
\paragraph{DEAL-WS.}
We use company-level workspace files from Workspace-Bench~\citep{tang2026workspacebench10benchmarkingai} (1,470 files across 5 worker roles) as source materials. 
We build its ability pool from workspace-based benchmarks~\citep{tang2026workspacebench10benchmarkingai,bai2026clawgymscalableframeworkbuilding, zai2026zclawbench, qwenclawbench, ding2026wildclawbench}. 
We combine these abilities with the workspace files and dependencies to construct 2,000 online tasks.
Figure~\ref{fig:deal-demo}(b) shows an example task in DEAL-WS, where the agent locates and inspects retention spreadsheets, runs Python analysis scripts, and produces a formatted workbook (\texttt{cohort\_q1.xlsx}).

For each scenario, we additionally filter 200 high-quality tasks for held-out evaluation.  
We provide a detailed overview of DEAL-WS and DEAL-RS in Appendix~\ref{app:data-details}.

\subsection{Execution, Feedback, and Evaluation Protocol}
\label{sec:protocol}

In DEAL, all methods use \texttt{nanobot} harness~\citep{hkuds2026nanobot} to execute tasks and record trajectories and artifacts.
We employ a user agent equipped with task rubrics to evaluate the delivered work and simulate user feedback.
Before scoring, a deliverable check verifies whether the required artifacts are generated in valid formats. Submissions without valid deliverables directly receive zero reward, acting as a multiplicative gate to prevent reward hacking (details provided in Appendix~\ref{app:reward-gate-details}), following recent practices in~\citep{mimo2026v26}.
For ease of analysis, we structure this feedback as \(F_t=(u_t,c_t)\), where \(u_t\in[0,1]\) is the utility score and \(c_t\) is textual feedback.
Both are returned once per task, while the rubrics and reference answers remain hidden.
This feedback protocol is an idealized simplification for fair comparison.
We compute online gains from live outcomes using Equation~\eqref{eq:objective}.
Held-out evaluation uses the final agent state after the full task stream.

\subsection{Implemented Baselines}
\label{sec:baselines}
We implement both training-free and parameter-based learning methods in DEAL. 
The training-free baselines use past experience to guide later tasks without updating model parameters. 
CASCADE retrieves successful cases as examples~\citep{guo2026cascade}, while EvoSkill turns feedback into reusable skills~\citep{alzubi2026evoskill}. 
For parameter-based learning, we use REINFORCE++~\citep{hu2025reinforcepp} to update the model from live trajectories and scalar utilities.

We further study parameter-based learning, as recent work shows its potential to internalize experience and enhance ability~\citep{harrington2026doescontinuallearningrequire,xia2026metaclawjusttalk}.
However, a single rollout may leave better behaviors undiscovered, even when its outcome is accurately evaluated.
In the next section, we analyze this exploration bottleneck and introduce our method, Self-Rewarded Exploration.

\section{SRE: Feedback-conditioned Self-rewarded Exploration}
\label{sec:method}

In this section, we first discuss the exploration bottleneck of single-rollout RL in Section~\ref{sec:why-expand}, motivating us to introduce SRE for learning from additional rollouts with imperfect self-rewards in Section~\ref{sec:sre-method}.
In Section~\ref{sec:theoretical-justification}, we provide theoretical support for SRE.

\subsection{Exploration Bottleneck of Single-rollout RL}
\label{sec:why-expand}

An intuitive approach to parameter-based learning is single-rollout RL~\citep{hu2025reinforcepp,hou2026sao,hu2026flashreinforce}.
It has shown success in gym-based agentic training before deployment, as it avoids group synchronization and supports asynchronous updates~\citep{hou2026sao}.
These environments still allow repeated attempts and feedback on the same task, providing further opportunities to discover different behaviors~\citep{xu2026singlestream}.

However, such opportunities cannot be assumed in deployment, as users may move on after one outcome feedback.
Whether the observed result is good or poor, alternative behaviors may remain unexplored. 
We refer to this limitation as an \textit{exploration bottleneck}.
A learned value model can support advantage estimation, but value pretraining and critic updates require careful configuration and add to the demands of local deployment~\citep{hou2026sao,justrl2,qi2026bestpracticecriticoptimization}.

We instead use additional rollouts to support both exploration and advantage estimation.
In deployment, self-evolution can run asynchronously with task execution~\citep{wang2026openclaw, xia2026metaclawjusttalk, tiwari2026learningfastslowllms}. 
From a saved task snapshot, the agent can generate additional offline rollouts in isolated copies when the inference server is free. 
Once these candidates receive cheap surrogate rewards, they can also provide signals to identify informative behaviors and estimate advantage.
This motivates Self-Rewarded Exploration (SRE), which views the live feedback as privileged information to assign rewards to the agent's extra rollouts.

\subsection{Learning with Additional Rollouts}
\label{sec:sre-method}

SRE consists of three components: offline rollout expansion, feedback-conditioned self-reward estimation, and Group-Relative Policy Optimization~\citep[GRPO;][]{shao2024deepseekmath}-style update.
The complete procedure is summarized in Appendix~\ref{app:pseudo-algo}.

\paragraph{Offline rollout expansion.}
For each task $x_t$, we save the initial environment state $s_t$ before live execution.
After receiving feedback $F_t$, we restore isolated copies of this snapshot and generate $K$ additional trajectories $\tilde\tau_t^{(k)}\sim\pi_{\btheta}(\cdot\mid x_t,s_t,M_{t-1})$, $k=1,\ldots,K$.
Here, each rollout uses the original task input and initial environment, and the live feedback is not added to the actor's input to keep the input conditions consistent with live execution~\citep{hubotter2026reinforcementlearningselfdistillation}.
Each trajectory is generated in the forked snapshot, without modifying the user's original workspace.

\paragraph{Feedback-conditioned self-reward estimation.}
We estimate surrogate rewards for these additional rollouts by having the frozen base model $\pi_{\btheta_0}$ score the actor's internal candidates.
Following recent work~\citep{selfdistilledreasoner2026,wang2026openclaw}, keeping both self-rubric and self-scoring agents fixed to $\pi_{\btheta_0}$ prevents reward drift and stabilizes RL.
The procedure has two stages.
First, a self-rubric agent takes the task $x_t$, the live trajectory $\tau_t$ with its artifacts, and the feedback $F_t=(u_t,c_t)$ as an anchor to build a shared rubric $\mathcal Q_t$, capturing unmet requirements, errors, and user preferences revealed on the live attempt.
Second, a self-scoring agent evaluates each candidate's execution and artifacts against $\mathcal Q_t$ without seeing $F_t$, so rubric imperfections act only as a shared group offset.
We query three evaluations per candidate and assign $\hat r_t^{(k)}$ as the median valid score, masking invalid cases (details provided in Appendix~\ref{app:experimental-settings}).
The resulting score thus reflects both task requirements and user feedback preferences.

\paragraph{GRPO-style update.}
Having the surrogate self-rewards in hand, we use a GRPO-style update to train the model~\citep{shao2024deepseekmath}. 
In implementation, we exclude the live user reward when calculating group-level advantages, as external scores exhibit a systematic offset from self-rewards.

\noindent
\begin{minipage}[t]{0.65\linewidth}
    \vspace{-0.6em}
    Figure~\ref{fig:reward-bias} shows that self-rewards tend to be higher than ground-truth rewards on the evaluated Gemma4-E4B-IT trajectories.
    Mixing them with the live reward can shift the group mean and distort the advantages.
    We therefore use only the $K$ offline trajectories and their self-rewards in each rollout group.
\par\vspace{0.25em}
\noindent
    Formally, for each task, let $\bar{\hat r}_t=K^{-1}\sum_{k=1}^{K}\hat r_t^{(k)}$ denote the mean self-reward.
    We compute the group-relative advantages as $\hat A_t^{(k)}=(\hat r_t^{(k)}-\bar{\hat r}_t)/[\operatorname{Std}(\{\hat r_t^{(j)}\}_{j=1}^{K})+\varepsilon]$, where $\varepsilon$ is a small constant for numerical stability.
\par\vspace{0.25em}
\noindent
    We apply a clipped GRPO-style policy update using these advantages, where only agent-generated tokens contribute to the policy loss, and tool observations serve as context.
\end{minipage}\hfill%
\begin{minipage}[t]{0.33\linewidth}
    \vspace{-0.45em}
    \centering
    \includegraphics[width=0.88\linewidth]{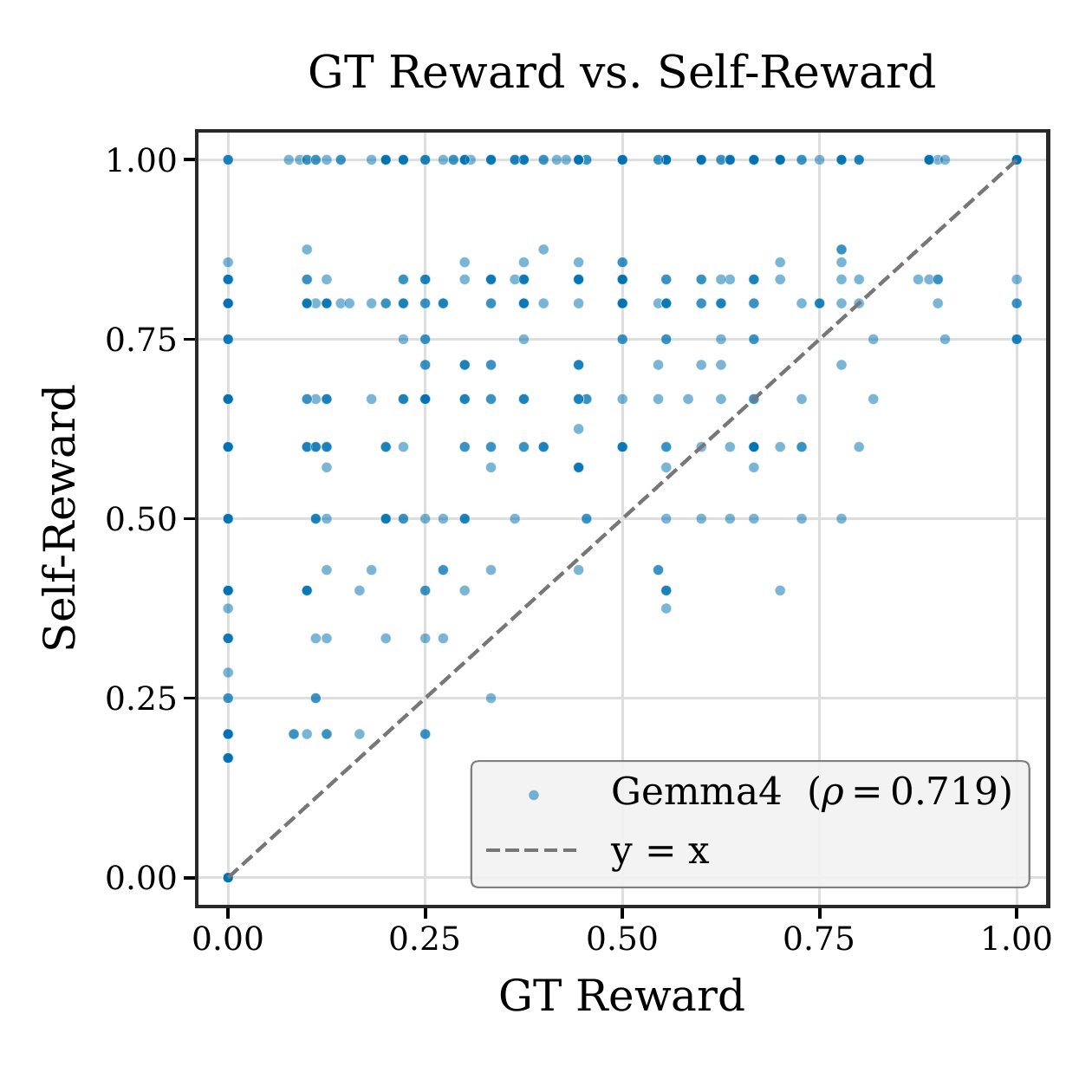}
    \vspace{-1.5em}
    \captionsetup{type=figure,hypcap=false}
    \captionof{figure}{
    GT vs.\ self-rewards for Gemma4 on \mbox{DEAL-RS}.
}
    \label{fig:reward-bias}
\end{minipage}

\subsection{Theoretical Justification: A Multi-armed Bandit View}
\label{sec:theoretical-justification}

In this subsection, we analyze rollout expansion with biased self-rewards from a Multi-Armed Bandit (MAB) perspective.
For a fixed task $x$ and a fixed policy $\pi_{\btheta}$, we view each complete trajectory as an arm with return $R_x(\tau)=u_x(\tau)\in[0,1]$.
Let $\mathcal T_K$ contain $K$ internal trajectories sampled independently from this policy, and let $R_x^*=\sup_\tau R_x(\tau)$.
Let $\hat R_x(\tau)$ denote its self-reward.
Considering the candidate selected by self-reward, i.e., $\hat\tau_K\in\arg\max_{\tau\in\mathcal T_K}\hat R_x(\tau)$, its expected simple regret can be decomposed as
\begin{equation*}
    \begin{aligned}
        \textsc{Reg}_K(x)
        &=
        R_x^*-\mathbb E[R_x(\hat\tau_K)]=
        \underbrace{
            R_x^*-
            \mathbb E\!\left[
                \max_{\tau\in\mathcal T_K}R_x(\tau)
            \right]
        }_{C_K(x):\ \text{coverage gap}}+
        \underbrace{
            \mathbb E\!\left[
                \max_{\tau\in\mathcal T_K}R_x(\tau)
                -
                R_x(\hat\tau_K)
            \right]
        }_{S_K(x):\ \text{selection error}}.
    \end{aligned}
\end{equation*}
The first term \(C_K(x)\) measures the gap from failing to discover higher-return trajectories. 
The second term \(S_K(x)\) measures the loss when self-rewards fail to select the best available candidate. 
For a single rollout, there is no candidate selection, so \(S_1(x)=0\) and \(\textsc{Reg}_1(x)=C_1(x)\). 
Increasing \(K\) can reduce \(C_K(x)\) through broader exploration, but self-reward errors may offset this gain. 
The following proposition bounds this selection error.

\begin{proposition}[Exploration with biased rewards]
    \label{prop:biased-reward-exploration}
    Suppose the estimated returns satisfy
    \begin{equation*}
        \hat R_x(\tau)
        =
        R_x(\tau)+b_x+e_x(\tau),
        \qquad
        |e_x(\tau)|\le\epsilon_r
    \end{equation*}
    for every candidate almost surely, where $b_x$ is shared across the group.
    Then
    \begin{equation*}
        \begin{aligned}
            S_K(x)&\le2\epsilon_r,\\
            \textsc{Reg}_1(x)-\textsc{Reg}_K(x)
            &\ge C_1(x)-C_K(x)-2\epsilon_r.
        \end{aligned}
    \end{equation*}
    Thus, expanded rollouts achieve lower expected selection regret when $C_1(x)-C_K(x)>2\epsilon_r$.
\end{proposition}

With Proposition~\ref{prop:biased-reward-exploration}, we can compare group-centered gradients from self-rewards and true returns.

\begin{corollary}[Group-relative gradient error]
    \label{cor:group-centered-gradient}
    Under the assumptions of Proposition~\ref{prop:biased-reward-exploration}, let $R_k=R_x(\tilde\tau^{(k)})$ and $\hat r_k=\hat R_x(\tilde\tau^{(k)})$, with group means $\bar R$ and $\bar{\hat r}$.
    Suppose $z_k=\nabla_{\btheta}\log\pi_{\btheta}(\tilde\tau^{(k)}\mid x)$ satisfies $\|z_k\|_2\le G$ almost surely.
    Treating rewards as fixed weights, define the GT and self-reward gradient estimates on the same candidate group as
    \begin{equation*}
        g_K
        =
        \frac{1}{K}\sum_{k=1}^{K}(R_k-\bar R)z_k,
        \qquad
        \hat g_K
        =
        \frac{1}{K}\sum_{k=1}^{K}(\hat r_k-\bar{\hat r})z_k.
        \label{eq:group-centered-gradients}
    \end{equation*}
    Then $\|\hat g_K-g_K\|_2\le2\epsilon_r G$ almost surely.
\end{corollary}

Both proofs are provided in Appendix~\ref{app:sre-proof}.
Proposition~\ref{prop:biased-reward-exploration} shows that exploration yields better candidates in expectation when coverage gains exceed selection errors.
Corollary~\ref{cor:group-centered-gradient} shows that small scoring errors keep group-centered gradients close to their GT counterparts.
Together, these results support learning from additional rollouts with feedback-conditioned self-rewards.

\section{Experiments}
\label{sec:experiments}

In this section, we conduct experiments on DEAL to study how agents learn from deployment experience.
We compare different adaptation methods and examine the effectiveness of SRE.

\subsection{Experimental Settings}
\label{sec:experimental-settings}

\paragraph{Evaluation settings.}
We evaluate online utility gains and held-out performance after deployment on DEAL-RS and DEAL-WS.
We also use AARRI-Bench~\citep{wang2026act} for DEAL-RS, and PinchBench~\citep{pinchbench2026} and ClawGym-Bench~\citep{bai2026clawgymscalableframeworkbuilding} for DEAL-WS.
Users also use these assistants for everyday conversations, so task improvements should not come at the cost of general capabilities.
We therefore evaluate MMLU-Pro~\citep{wang2024mmlupro}, BBEH~\citep{kazemi2025bbeh}, GPQA-D~\citep{rein2023gpqa}, AIME 2026~\citep{balunovic2025matharena}, MATH-500~\citep{lightman2023letsverify}, HumanEval+~\citep{liu2023evalplus}, and IFBench~\citep{pyatkin2025generalizing}.
These tests are run without harnesses, and their mean score is reported as \textit{General Capabilities}.
Online gain follows Equation~\eqref{eq:objective}; benchmark metrics and settings are given in Appendix~\ref{app:experimental-settings}.

\paragraph{Baselines and training.}
We use Gemma4-E4B-IT~\citep{gemmateam2026gemma4} for the main experiments.
We compare SRE with the frozen base agent, CASCADE~\citep{guo2026cascade}, EvoSkill~\citep{alzubi2026evoskill}, and REINFORCE++~\citep{hu2025reinforcepp}, using the implementations in Section~\ref{sec:baselines}.
Within each domain, methods share the task order and feedback interface.
SRE uses $K=7$ internal rollouts per task, with rollout generation and model updates running asynchronously.
Full configurations are given in Appendix~\ref{app:experimental-settings}.

\subsection{Main Results}
\label{sec:main-results}
\label{sec:exp-results}

\begin{table*}[!t]
    \centering
    \caption{
        Results for Gemma4-E4B-IT on DEAL-RS and DEAL-WS.
        Online gains are relative to the base agent as in Equation~\eqref{eq:objective}.
        Task abbreviations are defined in Appendix~\ref{app:task-categories}.
        Frozen models share the base model's general-capability scores (``--'').
        Best and second-best results within each domain are \textbf{bold} and \underline{underlined}, respectively.
    }
    \label{tab:main-results}

    \scriptsize
    \setlength{\tabcolsep}{2.8pt}
    \renewcommand{\arraystretch}{1.12}
    \vspace{0.3em}
    \resizebox{\textwidth}{!}{%
    \begin{tabular}{lcccccccccc}
        \toprule
        \multirow{2}{*}{\textbf{Method}}
        & \multirow{2}{*}{\shortstack{\textbf{Online}\\\textbf{Gain}}}
        & \multicolumn{6}{c}{\textbf{DEAL-RS (ID)}}
        & \multicolumn{2}{c}{\textbf{OOD Evaluation}}
        & \multirow{2}{*}{\shortstack{\textbf{General}\\\textbf{Capabilities}}} \\
        \cmidrule(lr){3-8}
        \cmidrule(lr){9-10}
        & & \textbf{Comp.}
          & \textbf{Align.}
          & \textbf{Verify.}
          & \textbf{Workflow}
          & \textbf{Audit}
          & \textbf{Avg.}
          & \multicolumn{2}{c}{\textbf{AARRI-Bench}}
          & \\
        \midrule

        Base
        & 0.00
        & 16.68 & 27.98 & 33.86 & 14.09 & 18.39 & 21.30
        & \multicolumn{2}{c}{35.38}
        & 52.91 \\

        EvoSkill
        & -1.07
        & 26.04 & 29.90 & 34.61 & 19.71
        & 24.70 & 26.90
        & \multicolumn{2}{c}{30.88}
        & -- \\

        CASCADE
        & +2.49
        & 25.52 & 23.93 & 37.30 & 16.38 & 19.59 & 24.93
        & \multicolumn{2}{c}{35.25}
        & -- \\

        REINFORCE++
        & \underline{+5.63}
        & \underline{31.34} & \underline{43.40} & \underline{55.26}
        & \textbf{29.68} & \underline{29.64} & \underline{36.95}
        & \multicolumn{2}{c}{\underline{37.01}}
        & \underline{52.97} \\

        \rowcolor{methodrowgreen}
        \textbf{SRE (Ours)}
        & \textbf{+10.66}
        & \textbf{36.91} & \textbf{47.89} & \textbf{55.93}
        & \underline{28.65} & \textbf{41.63} & \textbf{41.22}
        & \multicolumn{2}{c}{\textbf{38.60}}
        & \textbf{55.16} \\

        \midrule

        \multirow{2}{*}{\textbf{Method}}
        & \multirow{2}{*}{\shortstack{\textbf{Online}\\\textbf{Gain}}}
        & \multicolumn{6}{c}{\textbf{DEAL-WS (ID)}}
        & \multicolumn{2}{c}{\textbf{OOD Evaluation}}
        & \multirow{2}{*}{\shortstack{\textbf{General}\\\textbf{Capabilities}}} \\
        \cmidrule(lr){3-8}
        \cmidrule(lr){9-10}
        & & \textbf{Ops.}
          & \textbf{Log.}
          & \textbf{Res.}
          & \textbf{Dev.}
          & \textbf{Prod.}
          & \textbf{Avg.}
          & \textbf{PinchBench}
          & \textbf{ClawGym-Bench}
          & \\
        \midrule

        Base
        & 0.00
        & 48.52 & 44.78 & 38.34 & 37.02 & 32.55 & 40.24
        & 55.11 & 46.61
        & \underline{52.91} \\

        EvoSkill
        & -6.50
        & 47.58 & 49.22 & 35.35 & 38.61 & 33.74 & 40.90
        & 50.87 & 46.86
        & -- \\

        CASCADE
        & \underline{+0.92}
        & \textbf{52.50} & \underline{51.08} & 32.70
        & \underline{54.84} & \underline{36.84} & \underline{45.59}
        & 56.18 & 46.26
        & -- \\

        REINFORCE++
        & -0.44
        & 42.32 & 44.27 & \underline{41.36}
        & \textbf{61.31} & 33.42 & 44.54
        & \underline{57.99} & \underline{47.44}
        & \textbf{52.98} \\

        \rowcolor{methodrowgreen}
        \textbf{SRE (Ours)}
        & \textbf{+4.54}
        & \underline{49.22} & \textbf{51.67} & \textbf{45.61}
        & 50.69 & \textbf{36.90} & \textbf{46.81}
        & \textbf{59.54} & \textbf{50.27}
        & 52.87 \\

        \bottomrule
    \end{tabular}%
    }
\end{table*}

\begin{figure*}[t]
    \centering
    \begin{minipage}[t]{0.49\textwidth}
        \vspace{0pt}
        \centering
        \includegraphics[width=\linewidth]{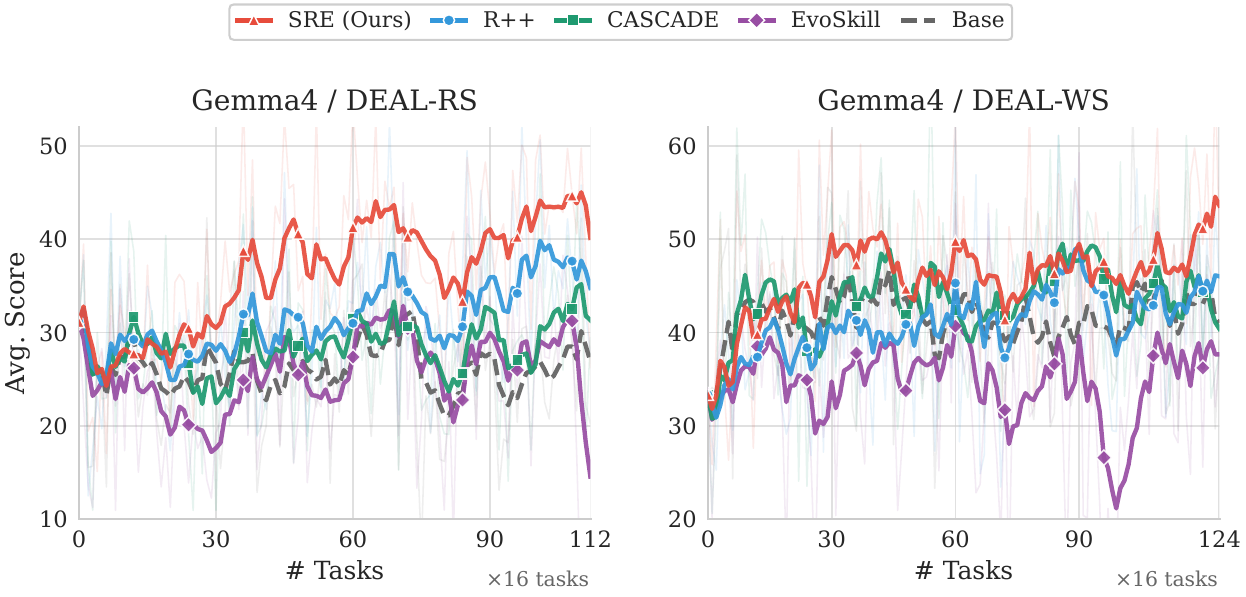}
        \vspace{-1.4em}
        \caption{
            Live task scores during deployment on DEAL-RS and DEAL-WS.
            Methods share the task order within each panel.
        }
        \label{fig:online-curves}
    \end{minipage}\hfill%
    \begin{minipage}[t]{0.49\textwidth}
        \vspace{0pt}
        \centering
        \includegraphics[width=\linewidth]{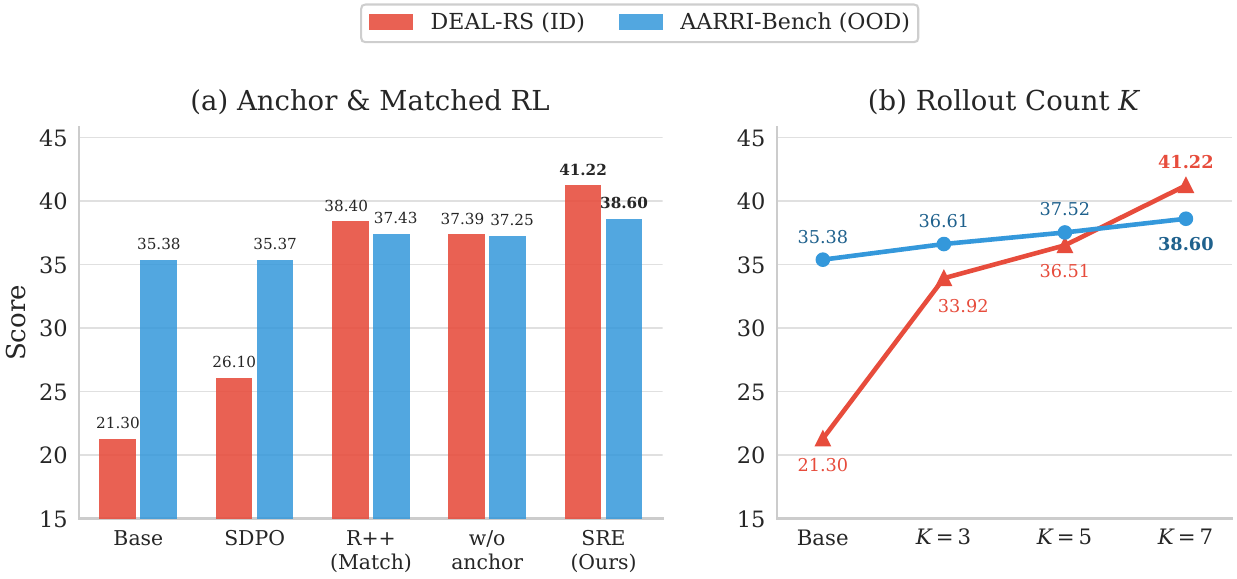}
        \vspace{-1.4em}
        \caption{
            End-to-end ablations on DEAL-RS: (a) anchor and matched RL; (b) rollout count $K \in \{3,5,7\}$.
        }
        \label{fig:sre-end-to-end-ablations}
    \end{minipage}
    \vspace{-0.4em}
\end{figure*}

Table~\ref{tab:main-results} and Figure~\ref{fig:online-curves} show three main findings.
First, SRE achieves the highest online utility gains on both deployment streams, outperforming REINFORCE++ and parameter-frozen baselines.
As shown in Figure~\ref{fig:online-curves}, memory retrieval and skill evolution plateau early or degrade when stored guidance mismatches the current task, whereas SRE steadily widens its margin over Base and single-rollout RL.
Second, on held-out ID tasks and external OOD benchmarks, SRE consistently attains the best overall performance in both domains, showing that deployment experience transfers to unseen tasks and evaluation harnesses.
Third, across the seven general-capability benchmarks, SRE preserves general capabilities without catastrophic forgetting, improving over Base on DEAL-RS and matching Base on DEAL-WS (per-benchmark results provided in Appendix~\ref{app:general-capability-details}).

\subsection{Analysis of Self-Rewarded Exploration}
\label{sec:sre-analysis}

\begin{figure*}[t]
    \centering
    \includegraphics[width=\textwidth]{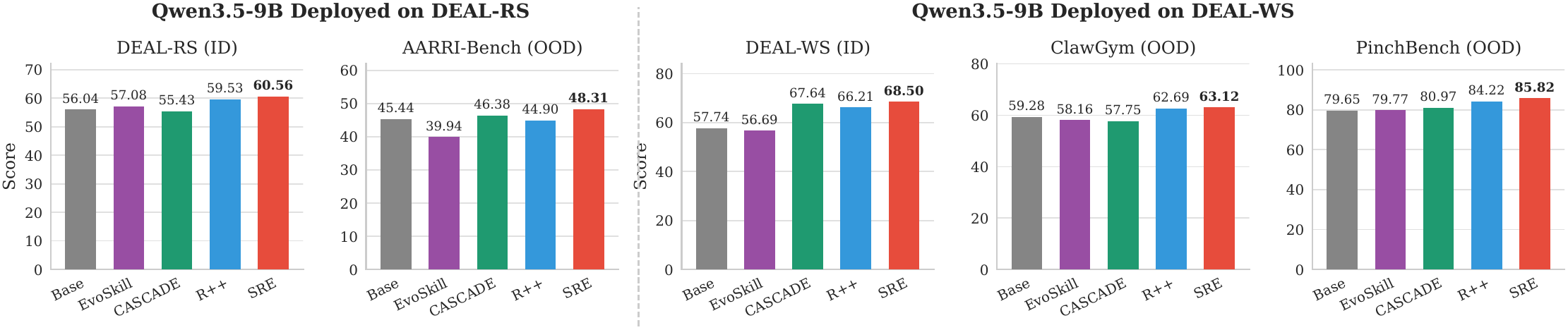}
    \vspace{-1.4em}
    \caption{
        Generalization to Qwen3.5-9B after deployment on \mbox{DEAL-RS} (left) and \mbox{DEAL-WS} (right).
        SRE achieves the highest ID and OOD scores across all five evaluation benchmarks.
    }
    \label{fig:cross-family-qwen}
    \vspace{-0.4em}
\end{figure*}

We next analyze the design choices, cross-family generalization, and practical robustness of SRE.

\paragraph{Role of the feedback anchor and compute-matched baselines.}
Figure~\ref{fig:sre-end-to-end-ablations}(a) compares SRE on DEAL-RS against three controlled baselines:
(1) \textbf{w/o anchor}, which builds rubrics from task descriptions alone over the same $K=7$ rollouts;
(2) \textbf{R++ (Match)}, which doubles REINFORCE++ gradient updates per step to match SRE's training compute (wall-clock breakdown provided in Appendix~\ref{app:compute-cost}); and
(3) \textbf{SDPO}~\citep{hubotter2026reinforcementlearningselfdistillation}, which applies feedback-conditioned token-level self-distillation on the same $K=7$ rollouts.
SRE outperforms all three on both ID and OOD evaluation: anchoring on live feedback improves within-group candidate ranking, while outcome-level rubric scoring proves more effective than extra single-rollout updates or token-level distillation.
Moreover, normalizing self-rewards internally avoids suppressing strong live trajectories when external and internal scores are mixed (diagnostics provided in Appendix~\ref{app:sre-reward-diagnostics}).

\paragraph{Effect of the rollout count.}
Figure~\ref{fig:sre-end-to-end-ablations}(b) evaluates policies trained end-to-end with $K \in \{3, 5, 7\}$ internal rollouts per task.
Both ID and OOD performance improve monotonically with $K$, as larger rollout groups discover higher-return candidates and stabilize group-relative advantage estimates.
Offline candidate-subset and gradient-alignment analyses confirm rising net gains that level off near $K=7$--$8$ (details provided in Appendix~\ref{app:candidate-gain-details}).

\paragraph{Cross-family generalization.}
We further evaluate SRE with Qwen3.5-9B~\citep{qwen3.5} on DEAL-RS and DEAL-WS, with results reported in Figure~\ref{fig:cross-family-qwen}.
Starting from a stronger base model than Gemma4, SRE still achieves the best performance across all five ID and OOD benchmarks, whereas CASCADE improves workspace ID scores at a slight OOD cost on ClawGym-Bench.
Controlled executor and skiller swaps with GLM-5.2 also indicate that parameter-frozen methods remain limited on small models by low skill-read rates and execution gaps even with inline skills (diagnostic analysis provided in Appendix~\ref{app:parameter-frozen-failures}).

\begin{figure*}[t]
    \centering
    \includegraphics[width=\textwidth]{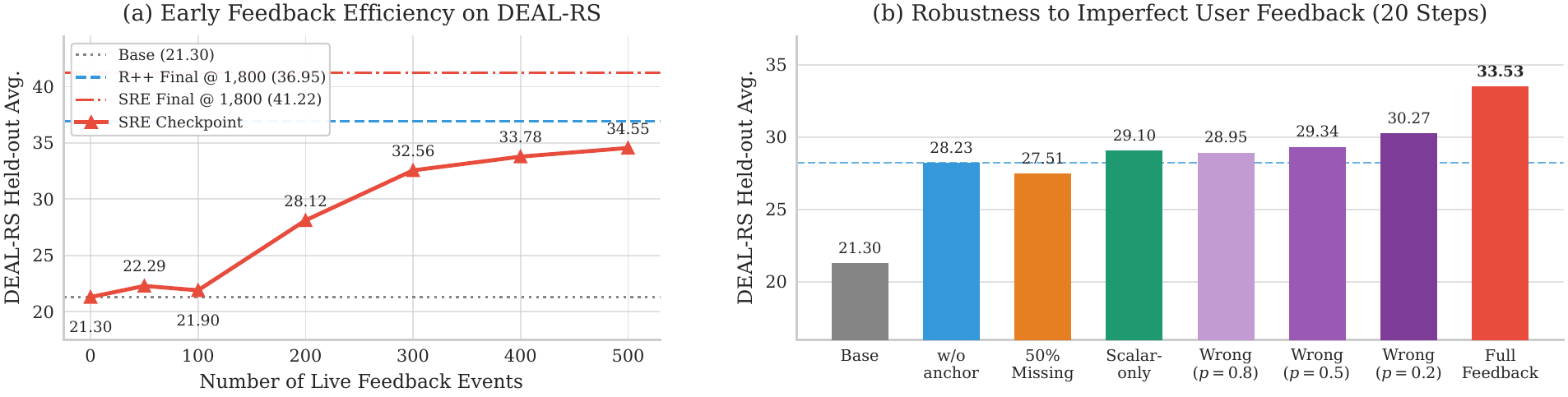}
    \vspace{-1.4em}
    \caption{
        Feedback efficiency and imperfect-feedback robustness of Gemma4-E4B-IT on DEAL-RS.
        (a) Held-out ID performance across early deployment checkpoints ($0$--$500$ feedbacks) versus full-stream policies ($1{,}800$ feedbacks).
        (b) Held-out ID performance after $20$ training steps under sparse ($50\%$ missing), scalar-only, and misleading ($p\in\{0.2, 0.5, 0.8\}$) user feedback.
    }
    \label{fig:fb-efficiency-robustness}
    \vspace{-0.4em}
\end{figure*}

\paragraph{Feedback efficiency and imperfect-feedback robustness.}
As shown in Figure~\ref{fig:fb-efficiency-robustness}(a), SRE improves rapidly after a brief warm-up on DEAL-RS, approaching full-stream REINFORCE++ within the first $500$ feedbacks.
Figure~\ref{fig:fb-efficiency-robustness}(b) further evaluates imperfect supervision over $20$ training steps, including $50\%$ missing feedback, scalar-only scores, and misleading feedback ($p\in\{0.2,0.5,0.8\}$).
SRE degrades gracefully and stays above Base and w/o anchor even at $p=0.8$, as $\mathcal Q_t$ remains grounded in the task and live artifacts while group centering cancels shared offsets $b_x$ following Proposition~\ref{prop:biased-reward-exploration}.
Seed-robustness results and qualitative cases are provided in Appendices~\ref{app:additional-rebuttal-ablations} and~\ref{app:case-studies}.

\section{Related Work}
\label{sec:related-work}

Prior benchmarks for post-deployment agent adaptation, such as DTLBench in CASCADE~\citep{guo2026cascade}, ClawArena~\citep{ji2026clawarenabenchmarkingaiagents}, and MetaClaw~\citep{xia2026metaclawjusttalk}, evaluate sequential task streams but largely center on QA-style queries rather than artifact-producing workflows, whereas ClawGym~\citep{bai2026clawgymscalableframeworkbuilding} scales executable tasks as isolated episodes.
In contrast, DEAL brings together cross-task relatedness, harness-native execution, and realistic user deliverables across workspace assistance and AI research under a single-feedback protocol.
Concurrent benchmarks such as Continual Learning Bench~\citep{asawa2026continuallearningbenchevaluating}, AgentCL~\citep{shu2026agentclrigorousevaluationcontinual}, and ServeLearnBench~\citep{zheng2026servelearnbench} similarly emphasize stateful environments and learning from serving experience, further underscoring these requirements for deployment-time evaluation.

Existing deployment-time adaptation methods retain experience through external memories, contexts, or reusable skills with frozen weights~\citep{chhikara2025mem0buildingproductionreadyai,zhang2026agenticcontextengineeringevolving,karten2026continualharnessonlineadaptation,alzubi2026evoskill,ma2026skillclawletskillsevolve}, or update model parameters through live interactions~\citep{wang2026openclaw,xia2026metaclawjusttalk,hebbar2026siaselfimprovingai}.
Single-rollout RL supports learning from individual trajectories and avoids group synchronization in asynchronous agentic training~\citep{hu2025reinforcepp,hou2026sao,hu2026flashreinforce,xu2026singlestream}.
Feedback-conditioned self-distillation offers another way to exploit textual feedback via token-level teacher supervision~\citep{hubotter2026reinforcementlearningselfdistillation,selfdistilledreasoner2026,shenfeld2026selfdistillationenablescontinuallearning}, yet often faces optimization instability in long-horizon agentic RL~\citep{realitycheck2026selfdistillation}.
SRE addresses the exploration bottleneck by expanding offline trajectories in isolated snapshots and scoring them with a feedback-anchored rubric, without requesting additional user evaluations.

\section{Discussion and Conclusion}
\label{sec:conclusion}

In this paper, we study deployment-time self-evolution under single-feedback interaction constraints and introduce DEAL, an executable testbed with related task streams in general workspace assistance and AI research.
To address the exploration bottleneck of single-rollout reinforcement learning, we propose Self-Rewarded Exploration (SRE), which expands offline rollouts in isolated environment snapshots and leverages the live trajectory and feedback as an anchor to construct shared self-rewards without additional user evaluations.
Our theoretical analysis characterizes when exploration gains outweigh self-reward errors and bounds the resulting error in group-centered policy gradients.
Experiments across Gemma4-E4B-IT and Qwen3.5-9B demonstrate that SRE consistently improves deployment-time utility, held-out execution, and out-of-distribution generalization across domains and imperfect feedback conditions, while preserving general capabilities.

\paragraph{Future work.}
Several promising directions remain open.
First, beyond full-parameter RL and external prompt memories, consolidating deployment experience into lightweight \textit{parametric memory}, such as online associative states with low-rank attention corrections in $\delta$-mem~\citep{lei2026deltamemefficientonlinememory} or context-to-weight consolidation in Cortex~\citep{zhang2026cortexconsolidatingcontextparametric}, offers an appealing route to internalize recurring procedures and user preferences while avoiding context bloat and general-capability interference.
Second, since not every feedback is worth updating on in practice, studying selective filtering to decide when to update alongside feedback-guided sampling, process-level credit assignment, and adaptive rollout budgets $K$ is an important direction.
Finally, evaluating self-evolving assistants on live human workflows while co-evolving general and domain-specific capabilities is an important next step.

\bibliographystyle{iclr2026_conference}
\bibliography{references}

@inproceedings{agrawal2026gepareflectivepromptevolution,
  author={Agrawal, Lakshya A. and Tan, Shangyin and Soylu, Dilara and Ziems, Noah and Khare, Rishi and Opsahl-Ong, Krista and Singhvi, Arnav and Shandilya, Herumb and Ryan, Michael J. and Jiang, Meng and Potts, Christopher and Sen, Koushik and Dimakis, Alexandros G. and Stoica, Ion and Klein, Dan and Zaharia, Matei and Khattab, Omar},
  title={{GEPA}: {R}eflective prompt evolution can outperform reinforcement learning},
  booktitle={Proceedings of the 14th International Conference on Learning Representations (ICLR'26)},
  address={Rio de Janeiro, Brazil},
  year={2026}
}

@inproceedings{alzubi2026evoskill,
  author={Alzubi, Salaheddin and Provenzano, Noah and Bingham, Jaydon and Calvo, Christian Alexander and Chen, Weiyuan and Vu, Tu},
  title={{EvoSkill}: {A}utomated skill discovery for multi-agent systems},
  booktitle={Proceedings of the 3rd Conference on Language Modeling (COLM'26)},
  address={San Francisco, CA},
  year={2026}
}

@misc{anthropic2026claudecode,
  title={{Claude Code}: {O}verview},
  author={{Anthropic}},
  howpublished={\url{https://code.claude.com/docs/en/overview}},
  year={2026}
}

@inproceedings{asawa2026continuallearningbenchevaluating,
  author={Asawa, Parth and Glaze, Christopher M. and Orlanski, Gabriel and Ramakrishnan, Ramya and Xu, Benji and Biswal, Asim and Chen, Vincent Sunn and Sala, Frederic and Zaharia, Matei and Gonzalez, Joseph E.},
  title={{Continual Learning Bench}: {E}valuating frontier {AI} systems in real-world stateful environments},
  booktitle={Advances in Neural Information Processing Systems 39 (NeurIPS'26)},
  address={Sydney, Australia},
  year={2026}
}

@article{bai2026clawgymscalableframeworkbuilding,
  title={{ClawGym}: {A} scalable framework for building effective {C}law agents},
  author={Bai, Fei and Song, Huatong and Sun, Shuang and Cheng, Daixuan and Yang, Yike and Hao, Chuan and Li, Renyuan and Chang, Feng and Wei, Yuan and Tao, Ran and Dai, Bryan and Yang, Jian and Zhao, Wayne Xin and Wen, Ji-Rong},
  journal={arXiv:2604.26904},
  year={2026}
}

@inproceedings{balunovic2025matharena,
  author={Balunovi{\'c}, Mislav and Dekoninck, Jasper and Petrov, Ivo and Jovanovi{\'c}, Nikola and Vechev, Martin},
  title={{MathArena}: {E}valuating {LLMs} on uncontaminated math competitions},
  booktitle={Advances in Neural Information Processing Systems 38 (NeurIPS'25)},
  address={San Diego, CA},
  year={2025}
}

@inproceedings{barres2026tau2bench,
  title={$\tau^2$-{Bench}: {E}valuating conversational agents in a dual-control environment},
  author={Barres, Victor and Dong, Honghua and Ray, Soham and Si, Xujie and Narasimhan, Karthik},
  booktitle={Proceedings of the 43rd International Conference on Machine Learning (ICML'26)},
  pages={6736--6782},
  address={Seoul, South Korea},
  year={2026}
}

@article{belcak2025smalllanguagemodelsfuture,
  title={Small language models are the future of agentic {AI}},
  author={Belcak, Peter and Heinrich, Greg and Diao, Shizhe and Fu, Yonggan and Dong, Xin and Muralidharan, Saurav and Lin, Yingyan Celine and Molchanov, Pavlo},
  journal={arXiv:2506.02153},
  year={2025}
}

@inproceedings{chan2025mlebench,
  title={{MLE-bench}: {E}valuating machine learning agents on machine learning engineering},
  author={Chan, Jun Shern and Chowdhury, Neil and Jaffe, Oliver and Aung, James and Sherburn, Dane and Mays, Evan and Starace, Giulio and Liu, Kevin and Maksin, Leon and Patwardhan, Tejal and Weng, Lilian and Madry, Aleksander},
  booktitle={Proceedings of the 13th International Conference on Learning Representations (ICLR'25)},
  address={Singapore},
  year={2025}
}

@inproceedings{chen2025scienceagentbench,
  title={{ScienceAgentBench}: {T}oward rigorous assessment of language agents for data-driven scientific discovery},
  author={Chen, Ziru and Chen, Shijie and Ning, Yuting and Zhang, Qianheng and Wang, Boshi and Yu, Botao and Li, Yifei and Liao, Zeyi and Wei, Chen and Lu, Zitong and Dey, Vishal and Xue, Mingyi and Baker, Frazier N. and Burns, Benjamin and Adu-Ampratwum, Daniel and Huang, Xuhui and Ning, Xia and Gao, Song and Su, Yu and Sun, Huan},
  booktitle={Proceedings of the 13th International Conference on Learning Representations (ICLR'25)},
  address={Singapore},
  year={2025}
}

@inproceedings{chhikara2025mem0buildingproductionreadyai,
  author={Chhikara, Prateek and Khant, Dev and Aryan, Saket and Singh, Taranjeet and Yadav, Deshraj},
  title={{Mem0}: {B}uilding production-ready {AI} agents with scalable long-term memory},
  booktitle={Proceedings of the 28th European Conference on Artificial Intelligence (ECAI'25)},
  pages={2993--3000},
  address={Bologna, Italy},
  year={2025}
}

@article{deepseekai2026deepseekv4highlyefficientmilliontoken,
  title={{DeepSeek-V4}: {T}owards highly efficient million-token context intelligence},
  author={{DeepSeek-AI}},
  journal={arXiv:2606.19348},
  year={2026}
}

@article{ding2026wildclawbench,
  title={{WildClawBench}: {A} benchmark for real-world, long-horizon agent evaluation},
  author={Ding, Shuangrui and Dai, Xuanlang and Xing, Long and Ding, Shengyuan and Liu, Ziyu and Yang, Jingyi and Yang, Penghui and Zhang, Zhixiong and Wei, Xilin and Fang, Xinyu and Ma, Yubo and Duan, Haodong and Shao, Jing and Wang, Jiaqi and Lin, Dahua and Chen, Kai and Zang, Yuhang},
  journal={arXiv:2605.10912},
  year={2026}
}

@article{gemmateam2026gemma4,
  title={{Gemma 4} technical report},
  author={{Gemma Team}},
  journal={arXiv:2607.02770},
  year={2026}
}

@article{glm5team2026glm5vibecodingagentic,
  title={{GLM-5}: {F}rom vibe coding to agentic engineering},
  author={{GLM-5-Team}},
  journal={arXiv:2602.15763},
  year={2026}
}

@article{guo2026cascade,
  title={{CASCADE}: {C}ase-based continual adaptation for large language models during deployment},
  author={Guo, Siyuan and Du, Yali and Chen, Hechang and Chang, Yi and Wang, Jun},
  journal={arXiv:2605.06702},
  year={2026}
}

@article{harrington2026doescontinuallearningrequire,
  title={When does continual learning require learning},
  author={Harrington, Anne and Saxena, Nayan and Murphy, Michael and Borovykh, Anastasia and Yun, Zeyu and Kamath, Sridhar and Kyi, Ara Eindra and Darrell, Trevor and Malik, Jitendra and Bai, Yutong},
  journal={arXiv:2607.07847},
  year={2026}
}

@article{hebbar2026siaselfimprovingai,
  title={{SIA}: {S}elf improving {AI} with harness \& weight updates},
  author={Hebbar, Prannay and Manawat, Yogendra and Verboomen, Samuel and Ivanova, Alesia and Palanimalai, Selvam and Bhatia, Kunal and Baskaran, Vignesh},
  journal={arXiv:2605.27276},
  year={2026}
}

@misc{hkuds2026nanobot,
  title={{nanobot}: {U}ltra-lightweight, open-source, self-hosted personal {AI} agent framework},
  author={{HKUDS}},
  howpublished={\url{https://github.com/HKUDS/nanobot}},
  year={2026}
}

@article{hou2026sao,
  title={Single-rollout asynchronous optimization for agentic reinforcement learning},
  author={Hou, Zhenyu and Li, Yujiang and Tang, Jie and Dong, Yuxiao},
  journal={arXiv:2607.07508},
  year={2026}
}

@article{hu2025reinforcepp,
  title={{REINFORCE++}: {S}tabilizing critic-free policy optimization with global advantage normalization},
  author={Hu, Jian and Liu, Jason Klein and Xu, Haotian and Shen, Wei},
  journal={arXiv:2501.03262},
  year={2025}
}

@article{hu2026flashreinforce,
  title={{FlashREINFORCE}: {C}ritic-free single-rollout asynchronous {RL} for agentic language models},
  author={Hu, Jian and Zhang, Yifan and Zhang, Hao and Xu, Binfeng and Zhang, Shaokun and Peng, Hongqing and Yu, Zhiding and Molchanov, Pavlo and Kautz, Jan and Dong, Yi},
  journal={arXiv:2611.18627},
  year={2026}
}

@article{hua2026cliuniverseverifiabletasksynthesis,
  title={{CLI-Universe}: {T}owards verifiable task synthesis engine for terminal agents},
  author={Hua, Zhanbo and Yao, Yifan and Xie, Weihao and Zhao, Yongchi and Liu, Minghao and Qiu, Ruizhi and Huang, Zhewei and Wang, Zun and Ji, Yiyan and Ye, Yunhai and Zhu, Letian and Lei, Xinping and Li, Han and Ma, Zhiyuan and Wang, Zili and Zhang, Zhaoxiang and Liu, Jiaheng},
  journal={arXiv:2606.22883},
  year={2026}
}

@inproceedings{hubotter2026reinforcementlearningselfdistillation,
  author={H{\"u}botter, Jonas and L{\"u}beck, Frederike and Behric, Lejs and Baumann, Anton and Bagatella, Marco and Marta, Daniel and Hakimi, Ido and Shenfeld, Idan and Kleine Buening, Thomas and Guestrin, Carlos and Krause, Andreas},
  title={Reinforcement learning via self-distillation},
  booktitle={Proceedings of the 43rd International Conference on Machine Learning (ICML'26)},
  pages={48677--48721},
  address={Seoul, South Korea},
  year={2026}
}

@article{ji2026clawarenabenchmarkingaiagents,
  title={{ClawArena}: {B}enchmarking {AI} agents in evolving information environments},
  author={Ji, Haonian and Xiong, Kaiwen and Han, Siwei and Xia, Peng and Qiu, Shi and Zhou, Yiyang and Liu, Jiaqi and Li, Jinlong and Li, Bingzhou and Zheng, Zeyu and Xie, Cihang and Yao, Huaxiu},
  journal={arXiv:2604.04202},
  year={2026}
}

@inproceedings{jimenez2024swebench,
  title={{SWE-bench}: {C}an language models resolve real-world {GitHub} issues?},
  author={Jimenez, Carlos E. and Yang, John and Wettig, Alexander and Yao, Shunyu and Pei, Kexin and Press, Ofir and Narasimhan, Karthik},
  booktitle={Proceedings of the 12th International Conference on Learning Representations (ICLR'24)},
  address={Vienna, Austria},
  year={2024}
}

@misc{justrl2,
  title={{JustRL II}: {S}caling small {LLMs} to {128K} reasoning with a critic},
  author={Pan, Haoxuan and Zhou, Chuyue and Li, Xin},
  howpublished={\url{https://panhaoxuan.notion.site/justrl-ii-scaling-small-llms-to-128k-reasoning-with-a-critic}},
  year={2026}
}

@article{karten2026continualharnessonlineadaptation,
  title={Continual harness: {O}nline adaptation for self-improving foundation agents},
  author={Karten, Seth and Zhang, Joel and Upaa, Jr., Tersoo and Feng, Ruirong and Li, Wenzhe and Shi, Chengshuai and Jin, Chi and Vodrahalli, Kiran},
  journal={arXiv:2605.09998},
  year={2026}
}

@inproceedings{kazemi2025bbeh,
  title={{BIG-Bench Extra Hard}},
  author={Kazemi, Mehran and Fatemi, Bahare and Bansal, Hritik and Palowitch, John and Anastasiou, Chrysovalantis and Mehta, Sanket Vaibhav and Jain, Lalit K. and Aglietti, Virginia and Jindal, Disha and Chen, Peter and Dikkala, Nishanth and Tyen, Gladys and Liu, Xin and Shalit, Uri and Chiappa, Silvia and Olszewska, Kate and Tay, Yi and Tran, Vinh Q. and Le, Quoc V. and Firat, Orhan},
  booktitle={Proceedings of the 63rd Annual Meeting of the Association for Computational Linguistics (ACL'25)},
  pages={26473--26501},
  address={Vienna, Austria},
  year={2025}
}

@article{kimiteam2026kimik3,
  title={{Kimi K3}: {O}pen frontier intelligence},
  author={{Kimi Team}},
  journal={arXiv:2607.24653},
  year={2026}
}

@inproceedings{lei2026deltamemefficientonlinememory,
  title={$\delta$-{Mem}: {E}fficient online memory for large language models},
  author={Lei, Jingdi and Zhang, Di and Li, Junxian and Wang, Weida and Fan, Kaixuan and Liu, Xiang and Liu, Qihan and Ma, Xiaoteng and Chen, Baian and Poria, Soujanya},
  booktitle={Advances in Neural Information Processing Systems 39 (NeurIPS'26)},
  address={Sydney, Australia},
  year={2026}
}

@article{li2026su01,
  title={Achieving gold-medal-level {Olympiad} reasoning via simple and unified scaling},
  author={Li, Yafu and Zhan, Runzhe and Zhang, Haoran and Zhang, Shunkai and Li, Yizhuo and Wang, Zhilin and Chen, Jiacheng and Wang, Futing and Hu, Xuyang and Fan, Yuchen and Xu, Bangjie and Su, Yucheng and Han, Xinmiao and Li, Chenxi and Lei, Haodi and Zhao, Yufeng and Lin, Zejin and Cheng, Qianjia and Zhu, Tong and Qu, Xiaoye and Cui, Ganqu and Ye, Peng and Luo, Yun and Lin, Zhouchen and Qiao, Yu and Zhou, Bowen and Ding, Ning and Cheng, Yu},
  journal={arXiv:2605.13301},
  year={2026}
}

@inproceedings{lightman2023letsverify,
  title={Let's verify step by step},
  author={Lightman, Hunter and Kosaraju, Vineet and Burda, Yura and Edwards, Harri and Baker, Bowen and Lee, Teddy and Leike, Jan and Schulman, John and Sutskever, Ilya and Cobbe, Karl},
  booktitle={Proceedings of the 12th International Conference on Learning Representations (ICLR'24)},
  address={Vienna, Austria},
  year={2024}
}

@inproceedings{liu2023evalplus,
  title={Is your code generated by {ChatGPT} really correct? {R}igorous evaluation of large language models for code generation},
  author={Liu, Jiawei and Xia, Chunqiu Steven and Wang, Yuyao and Zhang, Lingming},
  booktitle={Advances in Neural Information Processing Systems 36 (NeurIPS'23)},
  pages={21558--21572},
  address={New Orleans, LA},
  year={2023}
}

@article{ma2026skillclawletskillsevolve,
  title={{SkillClaw}: {L}et skills evolve collectively with agentic evolver},
  author={Ma, Ziyu and Yang, Shidong and Ji, Yuxiang and Wang, Xucong and Wang, Yong and Hu, Yiming and Huang, Tongwen and Chu, Xiangxiang},
  journal={arXiv:2604.08377},
  year={2026}
}

@inproceedings{merrill2026terminalbench,
  title={{Terminal-Bench}: {B}enchmarking agents on hard, realistic tasks in command line interfaces},
  author={Merrill, Mike A. and Shaw, Alexander G. and Carlini, Nicholas and Li, Boxuan and Raj, Harsh and Bercovich, Ivan and Shi, Lin and Shin, Jeong Yeon and Walshe, Thomas and Buchanan, E. Kelly and Shen, Junhong and Ye, Guanghao and Lin, Haowei and Poulos, Jason and Wang, Maoyu and Nezhurina, Marianna and Jitsev, Jenia and Lu, Di and Mastromichalakis, Orfeas Menis and Xu, Zhiwei and Chen, Zizhao and Liu, Yue and Zhang, Robert and Chen, Leon Liangyu and Kashyap, Anurag and Uslu, Jan-Lucas and Li, Jeffrey and Wu, Jianbo and Yan, Minghao and Bian, Song and Sharma, Vedang and Sun, Ke and Dillmann, Steven and Anand, Akshay and Lanpouthakoun, Andrew and Koopah, Bardia and Hu, Changran and Guha, Etash and Dreiman, Gabriel H. S. and Zhu, Jiacheng and Krauth, Karl and Zhong, Li and Muennighoff, Niklas and Amanfu, Robert and Tan, Shangyin and Pimpalgaonkar, Shreyas and Aggarwal, Tushar and Lin, Xiangning and Lan, Xin and Zhao, Xuandong and Liang, Yiqing and Wang, Yuanli and Wang, Zilong and Zhou, Changzhi and Heineman, David and Liu, Hange and Trivedi, Harsh and Yang, John and Lin, Junhong and Shetty, Manish and Yang, Michael and Omi, Nabil and Raoof, Negin and Li, Shanda and Zhuo, Terry Yue and Lin, Wuwei and Dai, Yiwei and Wang, Yuxin and Chai, Wenhao and Zhou, Shang and Wahdany, Dariush and She, Ziyu and Hu, Jiaming and Dong, Zhikang and Zhu, Yuxuan and Cui, Sasha and Saiyed, Ahson and Kolbeinsson, Arinbj{\"o}rn and Hu, Jesse and Rytting, Christopher Michael and Marten, Ryan and Wang, Yixin and Dimakis, Alex and Konwinski, Andy and Schmidt, Ludwig},
  booktitle={Proceedings of the 14th International Conference on Learning Representations (ICLR'26)},
  address={Rio de Janeiro, Brazil},
  year={2026}
}

@inproceedings{mialon2024gaia,
  title={{GAIA}: {A} benchmark for general {AI} assistants},
  author={Mialon, Gr{\'e}goire and Fourrier, Cl{\'e}mentine and Swift, Craig and Wolf, Thomas and LeCun, Yann and Scialom, Thomas},
  booktitle={Proceedings of the 12th International Conference on Learning Representations (ICLR'24)},
  address={Vienna, Austria},
  year={2024}
}

@article{mimo2026v26,
  title={{MiMo-V2.6}: {S}caling reinforcement learning towards self-improvement},
  author={{Xiaomi LLM-Core Team}},
  journal={arXiv:2610.11959},
  year={2026}
}

@article{minicpm4,
  title={{MiniCPM4}: {U}ltra-efficient {LLMs} on end devices},
  author={{MiniCPM Team}},
  journal={arXiv:2506.07900},
  year={2025}
}

@misc{nemo_rl,
  title={{NeMo RL}: {A} scalable and efficient post-training library},
  author={{NVIDIA}},
  howpublished={\url{https://github.com/NVIDIA-NeMo/RL}},
  year={2025}
}

@misc{openai2025codex,
  title={Introducing {Codex}},
  author={{OpenAI}},
  howpublished={\url{https://openai.com/index/introducing-codex/}},
  year={2025}
}

@misc{openai2026gpt55systemcard,
  title={{GPT-5.5} system card},
  author={{OpenAI}},
  howpublished={\url{https://openai.com/index/gpt-5-5-system-card/}},
  year={2026}
}

@misc{openai2026gpt56,
  title={{GPT-5.6}: {F}rontier intelligence that scales with your ambition},
  author={{OpenAI}},
  howpublished={\url{https://openai.com/index/gpt-5-6/}},
  year={2026}
}

@misc{openclaw2026,
  title={{OpenClaw}: {Y}our assistant, on your devices, in your chats},
  author={Steinberger, Peter and {OpenClaw Foundation} and {Contributors}},
  howpublished={\url{https://github.com/openclaw/openclaw}},
  year={2026}
}

@misc{opencode2026,
  title={{OpenCode}: {T}he open source coding agent},
  author={{Anomaly}},
  howpublished={\url{https://github.com/anomalyco/opencode}},
  year={2026}
}

@misc{pinchbench2026,
  title={{PinchBench}: {R}eal-world benchmarks for {AI} coding agents},
  author={{PinchBench}},
  howpublished={\url{https://github.com/pinchbench/skill}},
  year={2026}
}

@inproceedings{pmlr-v54-alquier17a,
  title={Regret bounds for lifelong learning},
  author={Alquier, Pierre and Mai, The Tien and Pontil, Massimiliano},
  booktitle={Proceedings of the 20th International Conference on Artificial Intelligence and Statistics (AISTATS'17)},
  pages={261--269},
  address={Fort Lauderdale, FL},
  year={2017}
}

@inproceedings{pyatkin2025generalizing,
  title={Generalizing verifiable instruction following},
  author={Pyatkin, Valentina and Malik, Saumya and Graf, Victoria and Ivison, Hamish and Huang, Shengyi and Dasigi, Pradeep and Lambert, Nathan and Hajishirzi, Hanna},
  booktitle={Advances in Neural Information Processing Systems 38 (NeurIPS'25)},
  address={San Diego, CA},
  year={2025}
}

@article{qi2026bestpracticecriticoptimization,
  title={Best practice critic optimization},
  author={Qi, Penghui and Zhou, Xiangxin and Lee, Wee Sun},
  journal={arXiv:2608.23566},
  year={2026}
}

@misc{qwen3.5,
  title={{Qwen3.5}: {T}owards native multimodal agents},
  author={{Qwen Team}},
  howpublished={\url{https://qwen.ai/blog?id=qwen3.5}},
  year={2026}
}

@misc{qwenclawbench,
  title={{QwenClawBench}: {R}eal-user-distribution benchmark for {OpenClaw} agents},
  author={{Qwen Team} and {Alibaba Data}},
  howpublished={\url{https://github.com/SKYLENAGE-AI/QwenClawBench}},
  year={2026}
}

@article{realitycheck2026selfdistillation,
  title={More methods, better agents? {A} reality check on self-distillation in agentic reinforcement learning},
  author={Zhong, Linhao and Fang, Bozhen and Wang, Xuehao and Shi, Duochao and Luo, Zekai and Yao, Guangming and Yuan, Yi and Chen, Hao and Chen, Jingdong and Shen, Chunhua},
  journal={arXiv:2610.19789},
  year={2026}
}

@inproceedings{rein2023gpqa,
  title={{GPQA}: {A} graduate-level {Google}-proof {Q\&A} benchmark},
  author={Rein, David and Hou, Betty Li and Stickland, Asa Cooper and Petty, Jackson and Pang, Richard Yuanzhe and Dirani, Julien and Michael, Julian and Bowman, Samuel R.},
  booktitle={Proceedings of the 1st Conference on Language Modeling (COLM'24)},
  address={Philadelphia, PA},
  year={2024}
}

@article{ren2026selfimprovementsmodernagenticsystems,
  title={Self-improvements in modern agentic systems: {A} survey},
  author={Ren, Zhe and Chen, Yimeng and Guo, Dandan and Rong, Guowei and Li, Tonghui and Xiong, R. B. and Lan, Qingfeng and Wang, Wenyi and {Li Nanbo} and Yang, Yibo and Zhuge, Mingchen and Schmidhuber, J{\"u}rgen},
  journal={arXiv:2607.13104},
  year={2026}
}

@inproceedings{selfdistilledreasoner2026,
  title={Self-distilled reasoner: {O}n-policy self-distillation for large language models},
  author={Zhao, Siyan and Xie, Zhihui and Liu, Mengchen and Huang, Jing and Pang, Guan and Chen, Feiyu and Grover, Aditya},
  booktitle={Proceedings of the 43rd International Conference on Machine Learning (ICML'26)},
  pages={162433--162448},
  address={Seoul, South Korea},
  year={2026}
}

@article{shao2024deepseekmath,
  title={{DeepSeekMath}: {P}ushing the limits of mathematical reasoning in open language models},
  author={Shao, Zhihong and Wang, Peiyi and Zhu, Qihao and Xu, Runxin and Song, Junxiao and Bi, Xiao and Zhang, Haowei and Zhang, Mingchuan and Li, Y. K. and Wu, Y. and Guo, Daya},
  journal={arXiv:2402.03300},
  year={2024}
}

@inproceedings{shenfeld2026selfdistillationenablescontinuallearning,
  title={Self-distillation enables continual learning},
  author={Shenfeld, Idan and Damani, Mehul and H{\"u}botter, Jonas and Agrawal, Pulkit},
  booktitle={Proceedings of the 43rd International Conference on Machine Learning (ICML'26)},
  pages={111542--111558},
  address={Seoul, South Korea},
  year={2026}
}

@article{shoeybi2019megatron,
  title={{Megatron-LM}: {T}raining multi-billion parameter language models using model parallelism},
  author={Shoeybi, Mohammad and Patwary, Mostofa and Puri, Raul and LeGresley, Patrick and Casper, Jared and Catanzaro, Bryan},
  journal={arXiv:1909.08053},
  year={2019}
}

@article{shu2026agentclrigorousevaluationcontinual,
  title={{AgentCL}: {T}oward rigorous evaluation of continual learning in language agents},
  author={Shu, Yiheng and Jim{\'e}nez Guti{\'e}rrez, Bernal and Jonnalagedda, Saisri Padmaja and Yao, Yuguang and Sun, Huan and Su, Yu},
  journal={arXiv:2606.02461},
  year={2026}
}

@misc{slime_github,
  title={{slime}: {A}n {LLM} post-training framework for {RL} scaling},
  author={Zhu, Zilin and Xie, Chengxing and Lv, Xin and {slime Contributors}},
  howpublished={\url{https://github.com/THUDM/slime}},
  year={2025}
}

@inproceedings{starace2025paperbench,
  title={{PaperBench}: {E}valuating {AI}'s ability to replicate {AI} research},
  author={Starace, Giulio and Jaffe, Oliver and Sherburn, Dane and Aung, James and Chan, Jun Shern and Maksin, Leon and Dias, Rachel and Mays, Evan and Kinsella, Benjamin and Thompson, Wyatt and Heidecke, Johannes and Glaese, Amelia and Patwardhan, Tejal},
  booktitle={Proceedings of the 42nd International Conference on Machine Learning (ICML'25)},
  pages={56843--56873},
  address={Vancouver, Canada},
  year={2025}
}

@inproceedings{tang2026workspacebench10benchmarkingai,
  title={{Workspace-Bench 1.0}: {B}enchmarking {AI} agents on workspace tasks with large-scale file dependencies},
  author={Tang, Zirui and Zhou, Xuanhe and Liu, Yumou and Li, Linchun and Wu, Yukai and Wang, Weizheng and Huang, Hongzhang and Zhou, Wei and Zhou, Jun and Song, Jiachen and Yu, Shaoli and Wang, Jinqi and Zhou, Zihang and Zhou, Hongyi and Lv, Yuting and Li, Jinyang and Liu, Jiashuo and Chen, Ruoyu and Liu, Chunwei and Li, Guoliang and Kang, Jihua and Wu, Fan},
  booktitle={Advances in Neural Information Processing Systems 39 (NeurIPS'26)},
  address={Sydney, Australia},
  year={2026}
}

@article{tiwari2026learningfastslowllms,
  title={Learning, fast and slow: {T}owards {LLMs} that adapt continually},
  author={Tiwari, Rishabh and Sareen, Kusha and Agrawal, Lakshya A. and Gonzalez, Joseph E. and Zaharia, Matei and Keutzer, Kurt and Dhillon, Inderjit S. and Agarwal, Rishabh and Khatri, Devvrit},
  journal={arXiv:2605.12484},
  year={2026}
}

@inproceedings{wang2024mmlupro,
  title={{MMLU-Pro}: {A} more robust and challenging multi-task language understanding benchmark},
  author={Wang, Yubo and Ma, Xueguang and Zhang, Ge and Ni, Yuansheng and Chandra, Abhranil and Guo, Shiguang and Ren, Weiming and Arulraj, Aaran and He, Xuan and Jiang, Ziyan and Li, Tianle and Ku, Max and Wang, Kai and Zhuang, Alex and Fan, Rongqi and Yue, Xiang and Chen, Wenhu},
  booktitle={Advances in Neural Information Processing Systems 37 (NeurIPS'24)},
  pages={95266--95290},
  address={Vancouver, Canada},
  year={2024}
}

@article{wang2026act,
  title={Act as a real researcher: {A} suite of benchmarks evaluating frontier {LLMs} and agentic harnesses in the research lifecycle},
  author={Wang, Jiayu and Lv, Weijiang and Fu, Bowen and Fu, Jing and Song, Jiayi and Zhang, Lingyu and Xue, Lanxuan and Chen, Luodi and Xin, Zepeng and Li, Kaiyu and Cao, Xiangyong},
  journal={arXiv:2606.07462},
  year={2026}
}

@article{wang2026openclaw,
  title={{OpenClaw-RL}: {T}rain any agent simply by talking},
  author={Wang, Yinjie and Chen, Xuyang and Jin, Xiaolong and Wang, Mengdi and Yang, Ling},
  journal={arXiv:2603.10165},
  year={2026}
}

@inproceedings{wu2025longmemeval,
  title={{LongMemEval}: {B}enchmarking chat assistants on long-term interactive memory},
  author={Wu, Di and Wang, Hongwei and Yu, Wenhao and Zhang, Yuwei and Chang, Kai-Wei and Yu, Dong},
  booktitle={Proceedings of the 13th International Conference on Learning Representations (ICLR'25)},
  address={Singapore},
  year={2025}
}

@article{xia2026metaclawjusttalk,
  title={{MetaClaw}: {J}ust talk -- {A}n agent that meta-learns and evolves in the wild},
  author={Xia, Peng and Chen, Jianwen and Yang, Xinyu and Tu, Haoqin and Liu, Jiaqi and Xiong, Kaiwen and Han, Siwei and Qiu, Shi and Ji, Haonian and Zhou, Yuyin and Zheng, Zeyu and Xie, Cihang and Yao, Huaxiu},
  journal={arXiv:2603.17187},
  year={2026}
}

@inproceedings{xie2024osworld,
  title={{OSWorld}: {B}enchmarking multimodal agents for open-ended tasks in real computer environments},
  author={Xie, Tianbao and Zhang, Danyang and Chen, Jixuan and Li, Xiaochuan and Zhao, Siheng and Cao, Ruisheng and Hua, Toh Jing and Cheng, Zhoujun and Shin, Dongchan and Lei, Fangyu and Liu, Yitao and Xu, Yiheng and Zhou, Shuyan and Savarese, Silvio and Xiong, Caiming and Zhong, Victor and Yu, Tao},
  booktitle={Advances in Neural Information Processing Systems 37 (NeurIPS'24)},
  pages={52040--52094},
  address={Vancouver, Canada},
  year={2024}
}

@article{xu2026polar,
  title={{Polar}: {A}gentic {RL} on any harness at scale},
  author={Xu, Binfeng and Zhang, Hao and Zhang, Shaokun and Han, Songyang and Liu, Mingjie and Hu, Jian and Diao, Shizhe and Jin, Zhenghui and Zou, Yunheng and Demoret, Michael and Kautz, Jan and Dong, Yi},
  journal={arXiv:2605.24220},
  year={2026}
}

@inproceedings{xu2026singlestream,
  title={Single-stream policy optimization},
  author={Xu, Zhongwen and Ding, Zihan},
  booktitle={Proceedings of the 14th International Conference on Learning Representations (ICLR'26)},
  address={Rio de Janeiro, Brazil},
  year={2026}
}

@article{yang2025qwen3technicalreport,
  title={{Qwen3} technical report},
  author={{Qwen Team}},
  journal={arXiv:2505.09388},
  year={2025}
}

@misc{zai2026zclawbench,
  title={{ZClawBench}: {A} benchmark for {OpenClaw}-style agent tasks},
  author={{Z.ai}},
  howpublished={\url{https://huggingface.co/datasets/zai-org/ZClawBench}},
  year={2026}
}

@inproceedings{zhang2026agenticcontextengineeringevolving,
  title={Agentic context engineering: {E}volving contexts for self-improving language models},
  author={Zhang, Qizheng and Hu, Changran and Upasani, Shubhangi and Ma, Boyuan and Hong, Fenglu and Kamanuru, Vamsidhar and Rainton, Jay and Wu, Chen and Ji, Mengmeng and Li, Hanchen and Thakker, Urmish and Zou, James and Olukotun, Kunle},
  booktitle={Proceedings of the 14th International Conference on Learning Representations (ICLR'26)},
  address={Rio de Janeiro, Brazil},
  year={2026}
}

@article{zhang2026cortexconsolidatingcontextparametric,
  title={{Cortex}: {C}onsolidating context into parametric memory},
  author={Zhang, Jintian and He, Feng and Liang, Xinnian and Xu, Ziwen and Xu, Haoming and Huang, Haojing and Wu, Shuangzhi and Zhang, Ningyu},
  journal={arXiv:2610.20399},
  year={2026}
}

@inproceedings{zhang2026pibenchevaluatingproactivepersonal,
  title={$\pi$-{Bench}: {E}valuating proactive personal assistant agents in long-horizon workflows},
  author={Zhang, Haoran and Xu, Luxin and Wang, Zhilin and Gui, Runquan and Zhang, Shunkai and Lei, Haodi and He, Zihao and He, Bingsu and Qin, Chicheng and Zhu, Tong and Qu, Xiaoye and Yang, Yang and Cheng, Yu and Li, Yafu},
  booktitle={Advances in Neural Information Processing Systems 39 (NeurIPS'26)},
  address={Sydney, Australia},
  year={2026}
}

@article{zheng2026servelearnbench,
  title={{ServeLearnBench}: {H}ow well can agents self-improve from serving experience?},
  author={Zheng, Haizhong and Di, Yizhuo and Sadhukhan, Ranajoy and Jin, Shuowei and Chen, Beidi},
  journal={arXiv:2610.07792},
  year={2026}
}

\clearpage
\appendix
\section*{Appendix}
\startcontents[appendix]
{\hypersetup{linkcolor=iclrdeepblue}\printcontents[appendix]{}{1}{\subsection*{Contents}}}

\section{Pseudocode of Self-Rewarded Exploration}
\label{app:pseudo-algo}

Algorithm~\ref{alg:sre} summarizes SRE under the single-feedback protocol.
The actor receives the same task information during live and internal execution.
Only the self-rubric stage accesses the live feedback, which guides candidate scoring through $\mathcal Q_t$.
We show the logical dependencies in order; internal rollout generation and policy updates can run asynchronously with later live tasks.

\begin{algorithm}[htbp]
\caption{Self-Rewarded Exploration (SRE)}
\label{alg:sre}
\begin{algorithmic}[1]
\Require Initial policy $\pi_{\btheta}$, task stream $(x_t)_{t=1}^{T}$, internal rollout count $K=7$, constant $\varepsilon>0$
\Ensure Updated policy and one delivered outcome per live task
\For{$t=1,\ldots,T$}
    \State Save the initial task environment $s_t$ and the available harness state $M_{t-1}$
    \State Execute $\tau_t\sim\pi_{\btheta}(\cdot\mid x_t,s_t,M_{t-1})$ in the live workspace
    \State Deliver its outcome and receive the single external feedback $F_t=(u_t,c_t)$
    \State Construct the shared rubric $\mathcal Q_t$ from $(x_t,\tau_t,F_t)$
    \For{$k=1,\ldots,K$}
        \State Restore an isolated copy of $s_t$
        \State Generate $\tilde\tau_t^{(k)}\sim\pi_{\btheta}(\cdot\mid x_t,s_t,M_{t-1})$ without $F_t$ in the actor input
        \State Score the candidate trace and artifacts using $\mathcal Q_t$ to obtain $\hat r_t^{(k)}$
    \EndFor
    \State Form the internal-only group $\{(\tilde\tau_t^{(k)},\hat r_t^{(k)})\}_{k=1}^{K}$
    \State Compute $\bar{\hat r}_t=K^{-1}\sum_{k=1}^{K}\hat r_t^{(k)}$ and $\sigma_t=\operatorname{Std}(\{\hat r_t^{(k)}\}_{k=1}^{K})$
    \State Set $\hat A_t^{(k)}=(\hat r_t^{(k)}-\bar{\hat r}_t)/(\sigma_t+\varepsilon)$ for each internal trajectory
    \State Update $\btheta$ with the clipped GRPO-style objective on agent-generated tokens
    \State Make the updated policy available for subsequent tasks
\EndFor
\end{algorithmic}
\end{algorithm}

The live trajectory provides the feedback anchor, but neither its tokens nor its GT reward enter the internal-only update group.
Tool observations provide context and are masked from the policy loss.
No internal candidate is submitted for another user evaluation.
Algorithm~\ref{alg:sre} describes a valid scored group; missing self-scores are not treated as observed GT labels.

\Needspace{16\baselineskip}
\section{Proofs for Section~\ref{sec:theoretical-justification}}
\label{app:sre-proof}

We use the notation introduced in Section~\ref{sec:theoretical-justification}.

\subsection{Proof of Proposition~\ref{prop:biased-reward-exploration}}
\begin{srepropappendix}[Exploration with biased rewards]
Suppose the estimated returns satisfy
    \begin{equation*}
        \hat R_x(\tau)
        =
        R_x(\tau)+b_x+e_x(\tau),
        \qquad
        |e_x(\tau)|\le\epsilon_r
    \end{equation*}
    for every candidate almost surely, where $b_x$ is shared across the group.
    Then
    \begin{equation*}
        \begin{aligned}
            S_K(x)&\le2\epsilon_r,\\
            \textsc{Reg}_1(x)-\textsc{Reg}_K(x)
            &\ge C_1(x)-C_K(x)-2\epsilon_r.
        \end{aligned}
        \end{equation*}
    Thus, expanded rollouts achieve lower expected selection regret when $C_1(x)-C_K(x)>2\epsilon_r$.

\end{srepropappendix}

\begin{proof}
Consider any realization of $\mathcal T_K$ on which the assumed error bounds hold.
Let
\[
    \tau_K^*
    \in
    \arg\max_{\tau\in\mathcal T_K}R_x(\tau),
    \qquad
    \hat\tau_K
    \in
    \arg\max_{\tau\in\mathcal T_K}\hat R_x(\tau).
\]
By the optimality of $\hat\tau_K$ under the estimated returns,
\[
    R_x(\hat\tau_K)+b_x+e_x(\hat\tau_K)
    \ge
    R_x(\tau_K^*)+b_x+e_x(\tau_K^*).
\]
The common offset $b_x$ cancels, giving
\[
    0
    \le
    R_x(\tau_K^*)-R_x(\hat\tau_K)
    \le
    e_x(\hat\tau_K)-e_x(\tau_K^*)
    \le
    2\epsilon_r.
\]
Taking expectations yields
\[
    S_K(x)
    =
    \mathbb E\!\left[
        R_x(\tau_K^*)-R_x(\hat\tau_K)
    \right]
    \le
    2\epsilon_r.
\]

For a single candidate, $S_1(x)=0$ and
$\textsc{Reg}_1(x)=C_1(x)$.
Applying the regret decomposition gives
\[
    \begin{aligned}
        \textsc{Reg}_1(x)-\textsc{Reg}_K(x)
        &=
        C_1(x)-C_K(x)-S_K(x)\\
        &\ge
        C_1(x)-C_K(x)-2\epsilon_r.
    \end{aligned}
\]
The right-hand side is positive whenever
$C_1(x)-C_K(x)>2\epsilon_r$, which proves the strict comparison.
\end{proof}

\paragraph{Remark.}
Under a fixed sampling policy, couple the candidate sets as $\mathcal T_{K+1}=\mathcal T_K\cup\{\tilde\tau^{(K+1)}\}$, where the additional trajectory is sampled independently from the same policy.
Then
\[
    C_K(x)-C_{K+1}(x)
    =
    \mathbb E\!\left[
        \left(
            R_x(\tilde\tau^{(K+1)})
            -
            \max_{\tau\in\mathcal T_K}R_x(\tau)
        \right)_+
    \right],
\]
where $(z)_+=\max\{z,0\}$.
Thus, the coverage gap decreases strictly whenever the additional rollout has a positive probability of outperforming the current best candidate.
The proposition specifies when this coverage gain exceeds the possible loss from reward estimation.

\subsection{Proof of Corollary~\ref{cor:group-centered-gradient}}
\label{app:group-centered-gradient-proof}

\begin{srecorappendix}[Group-relative gradient error]
Under the assumptions of Proposition~\ref{prop:biased-reward-exploration}, let $R_k=R_x(\tilde\tau^{(k)})$ and $\hat r_k=\hat R_x(\tilde\tau^{(k)})$, with group means $\bar R$ and $\bar{\hat r}$.
    Suppose $z_k=\nabla_{\btheta}\log\pi_{\btheta}(\tilde\tau^{(k)}\mid x)$ satisfies $\|z_k\|_2\le G$ almost surely.
    Treating rewards as fixed weights, define the GT and self-reward gradient estimates on the same candidate group as
    \begin{equation*}
        g_K
        =
        \frac{1}{K}\sum_{k=1}^{K}(R_k-\bar R)z_k,
        \qquad
        \hat g_K
        =
        \frac{1}{K}\sum_{k=1}^{K}(\hat r_k-\bar{\hat r})z_k.
        \end{equation*}
    Then $\|\hat g_K-g_K\|_2\le2\epsilon_r G$ almost surely.

\end{srecorappendix}

\begin{proof}
Fix a realization of the candidate group on which the reward-error and score-norm bounds hold.
Write $e_k=e_x(\tilde\tau^{(k)})$ and $\bar e=K^{-1}\sum_{k=1}^{K}e_k$.
Since $\hat r_k=R_k+b_x+e_k$, the group mean satisfies
$\bar{\hat r}=\bar R+b_x+\bar e$.
Hence,
\[
    (\hat r_k-\bar{\hat r})-(R_k-\bar R)
    =
    e_k-\bar e.
\]
The shared offset $b_x$ cancels.
Moreover, $|\bar e|\le\epsilon_r$, so
\[
    \left|
        (\hat r_k-\bar{\hat r})-(R_k-\bar R)
    \right|
    \le
    |e_k|+|\bar e|
    \le
    2\epsilon_r.
\]
Using the same trajectories and policy parameters for both gradient estimates gives
\[
    \begin{aligned}
        \|\hat g_K-g_K\|_2
        &=
        \left\|
            \frac{1}{K}\sum_{k=1}^{K}(e_k-\bar e)z_k
        \right\|_2\\
        &\le
        \frac{1}{K}\sum_{k=1}^{K}
        |e_k-\bar e|\,\|z_k\|_2\\
        &\le
        2\epsilon_r G.
    \end{aligned}
\]
This holds almost surely and completes the proof.
\end{proof}

\paragraph{Remark.}
The bound compares gradient estimates on the same candidate group, with rewards treated as fixed weights.
It allows biased and dependent scoring errors.
The result concerns the group-centered policy-gradient component without standard-deviation normalization or clipping, rather than the complete asynchronous GRPO update.

\section{DEAL Details}
\label{app:deal-details}

\subsection{Task Construction Details}
\label{app:task-construction}

Our construction pipeline combines grounded task synthesis with execution-based refinement~\citep{kimiteam2026kimik3,bai2026clawgymscalableframeworkbuilding,hua2026cliuniverseverifiabletasksynthesis}.
Figure~\ref{fig:data-construction} shows the three stages.
Each construction agent uses a harness to inspect files, call tools, and record its actions.

\paragraph{Task and evidence preparation.}
The resource curator organizes the source materials into an ability pool and an evidence pool.
The ability pool describes reusable operations, such as comparing benchmark results or reconciling a spreadsheet.
These abilities can be combined across tasks, so earlier experience can remain useful as the task context changes.
The evidence pool contains the files, code, assets, and dependencies needed to carry out those operations.
Before synthesis, we remove skills that require network access, unsupported multimodal inputs or outputs, or capabilities outside the two target domains.

\paragraph{Grounded task synthesis.}
The curator selects one or more abilities and defines a concrete work scenario.
It then retrieves the evidence needed for that scenario.
Given these inputs, the task generator writes the user request and its evaluation rubric together.
The request specifies the required work and deliverables, while the rubric checks requirements verifiable from the available evidence and final artifacts.
Hidden criteria and reference answers are excluded from actor-visible files.

\paragraph{Execution-based refinement and human review.}
The task tester executes each candidate and records its trajectory and artifacts.
The result is evaluated against the candidate rubric.
The task revisor inspects both the execution trace and the judging trace, then revises the request or rubric for another test.
If the evidence is missing or unsuitable, the case returns to the curator.
We also manually review the constructed tasks and rubrics for clear scope, sufficient evidence, feasibility, skill alignment, and verifiable evaluation criteria.
The review checks that task materials do not reveal the answer or prescribe the solution process.
Flagged cases are returned for revision.

\paragraph{Models and harnesses.}
The resource curator and task generator use GLM-5.2-FP8~\citep{glm5team2026glm5vibecodingagentic}.
For these two roles, DEAL-RS uses the nanobot harness~\citep{hkuds2026nanobot}, while DEAL-WS uses the Claude Code harness~\citep{anthropic2026claudecode}.
All task testers use nanobot.
The tester pool includes GLM-5.2-FP8, DeepSeek-V4-Flash~\citep{deepseekai2026deepseekv4highlyefficientmilliontoken}, Gemma4-31B-IT~\citep{gemmateam2026gemma4}, Qwen3.5-4B/9B~\citep{qwen3.5}, and Qwen3-4B-Thinking~\citep{yang2025qwen3technicalreport}.
The task revisor uses GPT-5.5 or GPT-5.6~\citep{openai2026gpt55systemcard,openai2026gpt56}.
All GPT-based construction agents run through the Codex harness~\citep{openai2025codex}.
All non-GPT models are self-hosted.
These construction harnesses are separate from the common nanobot executor used in our experiments.

\Needspace{12\baselineskip}
\subsection{Dataset Overview}
\label{app:data-details}
\label{app:task-categories}

\paragraph{DEAL-RS.}
We screen papers posted to arXiv's \texttt{cs.AI} category from September 2025 to June 2026 and retain 252 papers with associated open-source code repositories.
We build the ability pool from research skills collected from ClawHub\footnote{\url{https://clawhub.ai}} and SkillsHub\footnote{\url{https://www.skillhub.club/}}.
Combining these abilities with local paper and repository evidence gives 1,800 online tasks and 200 held-out tasks.
The five task families are benchmark comparison, method-to-code alignment, paper claim verification, research workflow package, and experiment design audit.
In Table~\ref{tab:main-results}, they are abbreviated as \textbf{Comp.}, \textbf{Align.}, \textbf{Verify.}, \textbf{Workflow}, and \textbf{Audit}, respectively.

\paragraph{DEAL-WS.}
We use the company-level files and dependencies in Workspace-Bench~\citep{tang2026workspacebench10benchmarkingai} as source materials.
The ability pool draws on Workspace-Bench, ClawGym~\citep{bai2026clawgymscalableframeworkbuilding}, ZClawBench~\citep{zai2026zclawbench}, QwenClawBench~\citep{qwenclawbench}, and WildClawBench~\citep{ding2026wildclawbench}.
We combine these abilities with grounded workspace materials to construct 2,000 online tasks and 200 held-out tasks.
The five role categories in Table~\ref{tab:main-results} are operations manager (\textbf{Ops.}), logistics manager (\textbf{Log.}), researcher (\textbf{Res.}), backend developer (\textbf{Dev.}), and product manager (\textbf{Prod.}).
For both domains, \textbf{Avg.} denotes the overall task score, \textbf{ID} denotes in-domain evaluation, and \textbf{OOD} denotes evaluation on external benchmarks.

\begin{figure}[t]
    \centering
    \includegraphics[width=\linewidth]{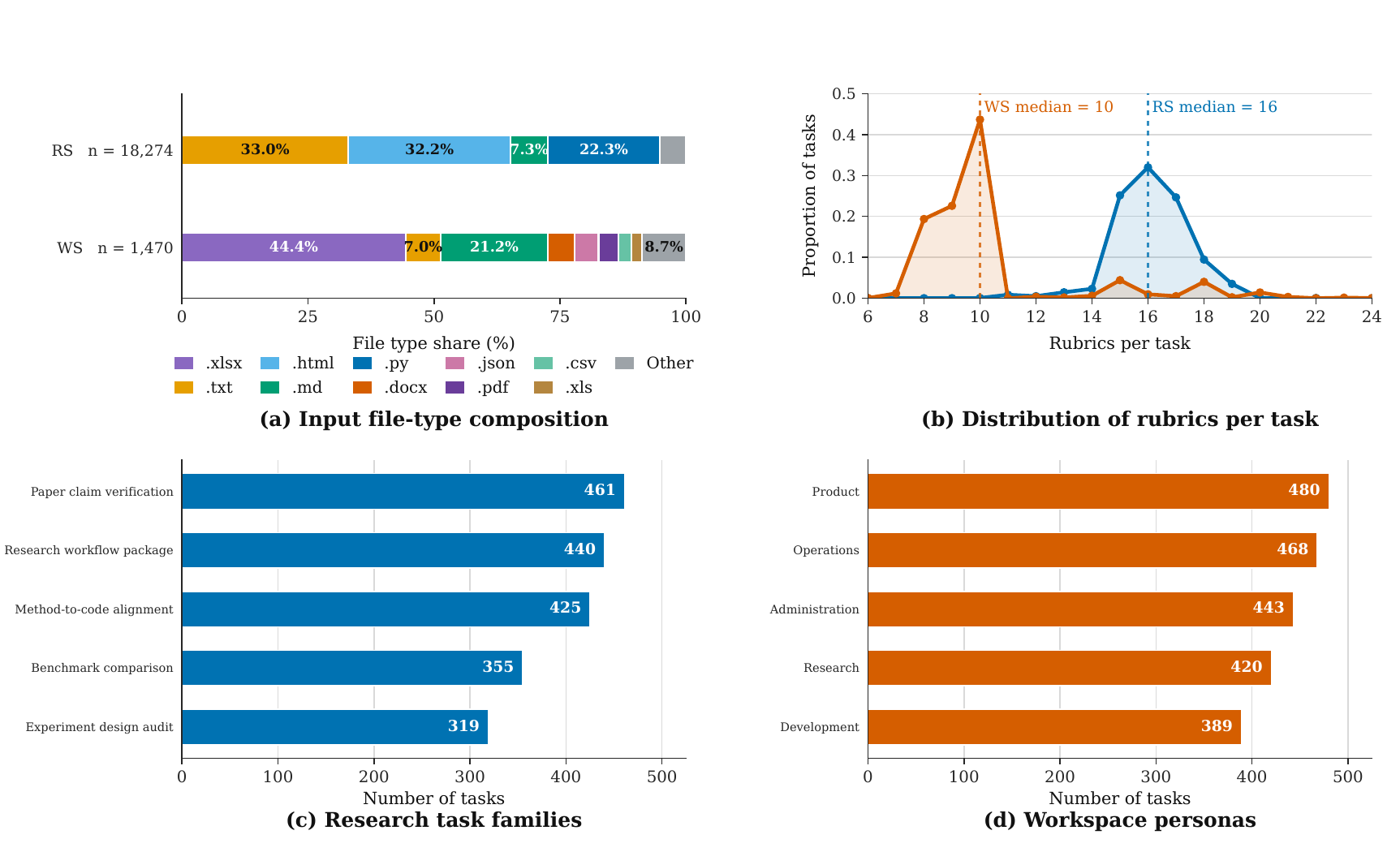}
    \caption{Dataset overview. (a) File-type shares among the inventoried input files in each domain; $n$ denotes files. (b) Distribution of the number of rubric criteria per task. (c) Research task-family counts across 1,800 online and 200 held-out tasks. (d) Workspace persona counts across the standard DEAL-WS benchmark of 2,000 online and 200 held-out tasks.}
    \label{fig:data-analysis}
\end{figure}

\paragraph{Dataset characteristics.}
Figure~\ref{fig:data-analysis}(a) summarizes 18,274 research input files and 1,470 workspace input files.
Research inputs mainly consist of text (33.0\%), HTML (32.2\%), and Python files (22.3\%).
Workspace inputs are dominated by spreadsheets (44.4\% \texttt{.xlsx}) and Markdown files (21.2\%).
These counts describe the input-file inventory, not the number of tasks.
Panel (b) shows that research tasks typically require more evaluation criteria: the median is 16, compared with 10 for workspace tasks.
Panel (c) contains 461 claim-verification tasks, 440 workflow tasks, 425 alignment tasks, 355 comparison tasks, and 319 audit tasks.
Panel (d) reports the workspace persona breakdown across all 2,000 online and 200 held-out tasks in DEAL-WS: product (480), operations (468), administration (443), research (420), and development (389).

\FloatBarrier
\section{Implementation and Experimental Details}
\label{app:implementation}

\subsection{Experimental Settings}
\label{app:experimental-settings}

\paragraph{Training infrastructure.}
We build our training infrastructure on slime~\citep{slime_github}.
Our asynchronous design draws on the infrastructure designs of
Polar~\citep{xu2026polar},
OpenClaw-RL~\citep{wang2026openclaw},
and ClawGym~\citep{bai2026clawgymscalableframeworkbuilding}.
We decouple rollout execution and reward computation from policy
optimization, allowing rollout collection and training to overlap.
For Gemma4-E4B-IT, we integrate the Megatron backend adapted in
NeMo RL~\citep{nemo_rl,shoeybi2019megatron} into this training pipeline.

\paragraph{Training settings and evaluator roles.}
We use Gemma4-E4B-IT~\citep{gemmateam2026gemma4} and Qwen3.5-9B~\citep{qwen3.5} as the deployed models.
Table~\ref{tab:training-settings} lists the rollout and reward settings for both domains.
Specifically, three distinct evaluator roles participate in our setup:
(1) the \textbf{external user simulator} uses Gemma4-31B-IT with access to the hidden ground-truth rubric to evaluate the single live execution and return $(u_t, c_t)$ as part of the testbed environment;
(2) the \textbf{self-rubric agent} uses the \emph{frozen initial model} $\pi_{\btheta_0}$ (Gemma4-E4B-IT or Qwen3.5-9B) to construct $\mathcal Q_t$ from $(x_t, \tau_t, F_t)$, where \texttt{ANCHOR.md} packages the live response, public feedback $c_t$, and scalar score $u_t$; and
(3) the \textbf{self-scoring agent} uses the same \emph{frozen initial model} $\pi_{\btheta_0}$ to score each offline rollout against $\mathcal Q_t$ without observing $F_t$ directly (prompt templates and schemas provided in Appendix~\ref{app:prompts}).
Keeping both internal evaluators frozen prevents reward drift during RL~\citep{selfdistilledreasoner2026,wang2026openclaw}.
SRE uses seven internal rollouts per task and forms advantage groups from their self-rewards only.
The self-scoring procedure runs three independent scoring sessions per candidate and takes the median valid score (or the mean when two sessions succeed); candidates with no valid score, or groups with invalid anchor or rubric outputs, are masked from advantage normalization and policy updates.
Rollout generation and model updates run asynchronously; tool observations are masked from the policy loss.

\par
\begin{table}[htbp]
    \centering
    \caption{Gemma4-E4B-IT rollout and reward configuration for DEAL-WS and DEAL-RS.}
    \label{tab:training-settings}
    \vspace{0.3em}
    \footnotesize
    \setlength{\tabcolsep}{4.8pt}
    \begin{tabular}{lrr}
        \toprule
        Setting & Workspace tasks & Research tasks \\
        \midrule
        Agent harness & nanobot & nanobot \\
        Context window & $64$K & $72$K \\
        Maximum output & $8$K & $16$K \\
        Temperature & $0.7$ & $0.7$ \\
        Top-$p$ & $1.0$ & $1.0$ \\
        Top-$k$ & $-1$ & $-1$ \\
        Maximum tool iterations & $200$ & $200$ \\
        \midrule
        External user simulator (GT judge) & Gemma4-31B-IT & Gemma4-31B-IT \\
        Self-rubric \& self-scoring agents & Frozen $\pi_{\btheta_0}$ (Gemma4-E4B-IT) & Frozen $\pi_{\btheta_0}$ (Gemma4-E4B-IT) \\
        Judge temperature & $0.7$ & $0.7$ \\
        Judge maximum output & $8$K & $8$K \\
        Judge timeout & $2{,}700$ s & $900$ s \\
        Agentic judge timeout & -- & $1{,}800$ s \\
        Maximum judge attempts & $3$ & $3$ \\
        \bottomrule
    \end{tabular}
\end{table}

\paragraph{Evaluation settings.}
For \textbf{evaluation on DEAL}, we evaluate each saved system state on the workspace and research evaluation sets, each containing $200$ tasks. Workspace evaluation uses a $64$K context window and an $8$K output limit, while research evaluation uses an $80$K context window and a $16$K output limit. The agent harness and tool configuration otherwise match those used during deployment. We keep the evaluated state fixed and do not learn from evaluation feedback.
\par\smallskip

\noindent
For \textbf{evaluation beyond DEAL}, we use ClawGym-Bench~\citep{bai2026clawgymscalableframeworkbuilding} and PinchBench~\citep{pinchbench2026} for workspace tasks, and AARRI-Bench~\citep{wang2026act} for research tasks.
We filter only tasks that are incompatible with the available execution environment.
For ClawGym-Bench, we remove 13 tasks that require network access and retain 187 tasks.
For PinchBench\footnote{\url{https://github.com/pinchbench/skill}}, we remove 20 tasks: 14 require live web access, external repositories, or video processing; two require visual understanding or generation; and four depend on an integrated \texttt{fws} and \texttt{gh/gws} environment.
The remaining split contains 127 tasks.
For AARRI-Bench\footnote{\url{https://github.com/AARR-bench/AARRI-bench}}, we exclude one task requiring web search and one requiring SVG evaluation, leaving 80 tasks.
For AARRI-Bench, we generate one evaluated response per task.
Table~\ref{tab:external-benchmark-settings} lists the shared evaluation budgets.
\par
\begin{table}[htbp]
    \centering
    \caption{Evaluation settings for external benchmarks.}
    \label{tab:external-benchmark-settings}
    \vspace{0.3em}
    \footnotesize
    \setlength{\tabcolsep}{7pt}
    \begin{tabular}{lcc}
        \toprule
        Setting & ClawGym-Bench / PinchBench & AARRI-Bench \\
        \midrule
        Agent harness & nanobot & nanobot \\
        Context window & $64$K & $64$K \\
        Maximum output & $8$K & $8$K \\
        Temperature & $0.7$ & $0.7$ \\
        Top-$p$ & $1.0$ & $1.0$ \\
        Top-$k$ & $-1$ & $-1$ \\
        Maximum task attempts & $3$ & $3$ \\
        Overall task timeout & $4{,}200$ s & $4{,}200$ s \\
        Agent execution timeout & $3{,}600$ s & $3{,}600$ s \\
        Evaluator & See text & pytest verifier \\
        Evaluator timeout & $300$ s & -- \\
        Maximum evaluator attempts & $3$ & -- \\
        \bottomrule
    \end{tabular}
\end{table}
\par\noindent
For ClawGym-Bench, we combine code-based checks with rubric-based assessment by Gemma4-31B-IT~\citep{gemmateam2026gemma4}.
The code checks evaluate mechanically verifiable requirements, while the rubric judge assesses the delivered work against the task criteria.
PinchBench also uses Gemma4-31B-IT for judging.
AARRI-Bench uses its deterministic \texttt{pytest} verifier.
Missing or failed tasks receive zero reward; we do not exclude them from benchmark averages.
\par\smallskip

\noindent\textbf{General capability evaluation.}
We evaluate MMLU-Pro~\citep{wang2024mmlupro}, BBEH~\citep{kazemi2025bbeh}, GPQA-D~\citep{rein2023gpqa}, AIME 2026~\citep{balunovic2025matharena}, MATH-500~\citep{lightman2023letsverify}, HumanEval+~\citep{liu2023evalplus}, and IFBench~\citep{pyatkin2025generalizing} without tools or web access. We enable reasoning for Gemma4-E4B-IT and use an $80$K context limit. For AIME 2026 and MATH-500, we sample four responses per problem and report average accuracy; all other benchmarks use one response per item. For MMLU-Pro and GPQA-D, we report accuracy. For BBEH, we report the harmonic mean across its 23 tasks. For HumanEval+, we report pass@$1$. For IFBench, we report prompt-level strict accuracy, and compute \textit{General Capabilities} as the unweighted arithmetic mean over the seven benchmarks.

Gemma4-E4B-IT uses a $16$K output limit, except on MATH-500, where the limit is $32$K.
The default decoding parameters are temperature $0.7$, top-$p$ $1.0$, and top-$k$ $-1$.
MATH-500 instead uses temperature $0.6$ and top-$p$ $0.95$.
We keep these settings fixed across base and adapted models.

\paragraph{Parameter-frozen baselines.}
For CASCADE~\citep{guo2026cascade}, we keep the retrieval encoder fixed as the memory grows during deployment.
For EvoSkill~\citep{alzubi2026evoskill}, we follow its original skill evolution procedure within our ordered deployment stream. Each update uses only the trajectories and feedback observed up to that point, and newly evolved skills become available only for subsequent tasks.
Concretely, we retain EvoSkill's official OpenCode-based skiller~\citep{opencode2026} and evolution logic, while replacing only its task-execution component with the same Nanobot executor harness used by the other methods.
The executor model remains fixed for both baselines.

\subsection{Deliverable Format Gating and Reward-Hacking Prevention}
\label{app:reward-gate-details}

In both workspace and research domains, tasks require concrete file deliverables such as spreadsheets, Word documents, or structured reports.
Without format verification, an agent may fail to construct valid files while still producing plausible text; for example, writing raw Markdown bytes under a \texttt{.docx} extension instead of generating a valid Word binary.
When evaluated on text content alone, a rubric-based user simulator can still award partial credit to such malformed outputs, creating a shortcut that policies increasingly exploit during RL.

\begin{figure}[htbp]
    \centering
    \includegraphics[width=0.88\linewidth]{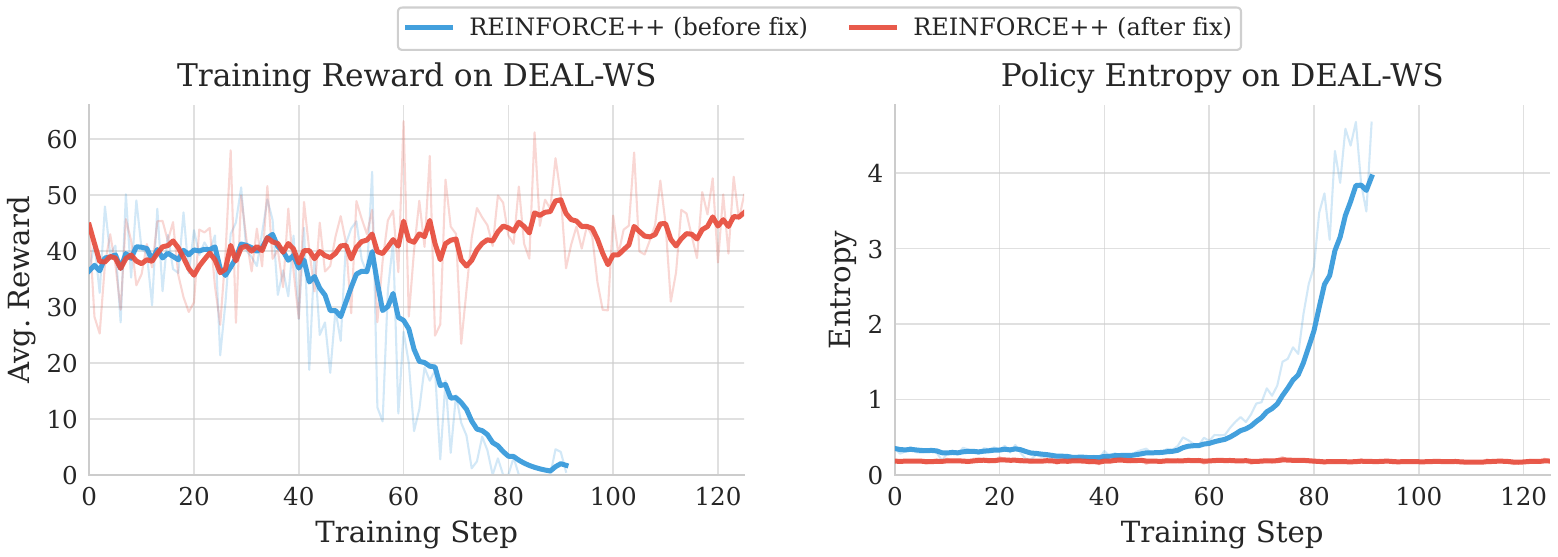}
    \caption{Training reward and policy entropy of Gemma4-E4B-IT with REINFORCE++ on DEAL-WS before and after enforcing the deliverable format check.}
    \label{fig:app-reward-gate}
\end{figure}

To prevent such reward hacking~\citep{mimo2026v26}, we apply a deterministic format gate prior to rubric scoring: submissions with missing or malformed deliverables directly receive zero reward, $R = \mathbb{I}[\text{valid deliverable}] \cdot R_{\text{rubric}}$.
Figure~\ref{fig:app-reward-gate} illustrates this effect for Gemma4-E4B-IT trained with REINFORCE++ on DEAL-WS.
Without the format gate, training reward collapses after step~40 accompanied by a sharp surge in policy entropy, whereas enforcing the gate stabilizes entropy and sustains steady reward improvement.

\subsection{Computational Cost Breakdown}
\label{app:compute-cost}

For self-hosted deployment, practical training overhead is governed by end-to-end wall-clock time and GPU-hours.
Table~\ref{tab:cost-breakdown} reports the stage breakdown of SRE ($K=7$) and REINFORCE++ over 113 training steps (1,800 tasks) on DEAL-RS using an $8\times\text{H200}$ node.
Each SRE trajectory averages $106.54\,\text{s}$ of active execution ($121.64\,\text{s}$ including queuing, initialization, and teardown), reflecting multi-turn tool execution rather than LLM decoding alone.
Because rollout collection and reward estimation run asynchronously alongside policy updates, much of the additional sampling latency is overlapped, resulting in $24.06\,\text{h}$ ($192.5$ GPU-hours) for SRE versus $18.02\,\text{h}$ ($144.2$ GPU-hours) for REINFORCE++ ($\sim 1.33\times$ wall-clock time).

\begin{table}[htbp]
    \centering
    \caption{Wall-clock stage breakdown over 113 training steps on DEAL-RS on an $8\times\text{H200}$ node. Since rollout collection and training overlap asynchronously, stage durations do not sum directly.}
    \label{tab:cost-breakdown}
    \vspace{0.3em}
    \footnotesize
    \setlength{\tabcolsep}{7pt}
    \begin{tabular}{lrrl}
        \toprule
        Stage ($8\times\text{H200}$) & SRE ($K=7$) & REINFORCE++ & Notes \\
        \midrule
        Actor update & $12.69$ h & $10.15$ h & Policy gradient optimization \\
        Log-prob forward & $3.09$ h & $2.47$ h & Forward pass for log-prob \\
        Data preprocessing & $1.65$ h & $1.32$ h & Training batch preparation \\
        Weight synchronization & $0.25$ h & $0.25$ h & Syncing weights to rollout workers \\
        Rollout \& reward wait & $12.94$ h & $7.51$ h & Overlapped with training \\
        \midrule
        \textbf{Total wall-clock} & \textbf{24.06 h} & \textbf{18.02 h} & \textbf{$192.5$ vs.\ $144.2$ GPU-hours ($\sim 1.33\times$)} \\
        \bottomrule
    \end{tabular}
\end{table}
\section{Additional Experiments}
\subsection{Feedback Anchoring and Reward-Source Mixing}
\label{sec:feedback-exp}
\label{app:sre-reward-diagnostics}
\label{app:reward-grouping-details}

We use 100 randomly sampled Gemma4 task groups from DEAL-RS to diagnose self-reward ranking quality and advantage estimation.
Each group has one live GT reward and seven internal trajectories with GT and self-reward pairs.
With the seven candidate trajectories and GT labels held fixed, we compare three self-reward conditions: \textit{w/ anchor} (conditioning the self-rubric on the live attempt and clean user feedback), \textit{w/o anchor} (task-only rubric without user feedback), and \textit{wrong feedback} (corrupted user feedback).
The recorded self-scores use the median of successful repeated evaluations; we retain recorded zeros and do not replace missing scores with zero.
All 100 randomly sampled groups have complete pairs.

\begin{figure}[htbp]
    \centering
    \includegraphics[width=0.92\linewidth]{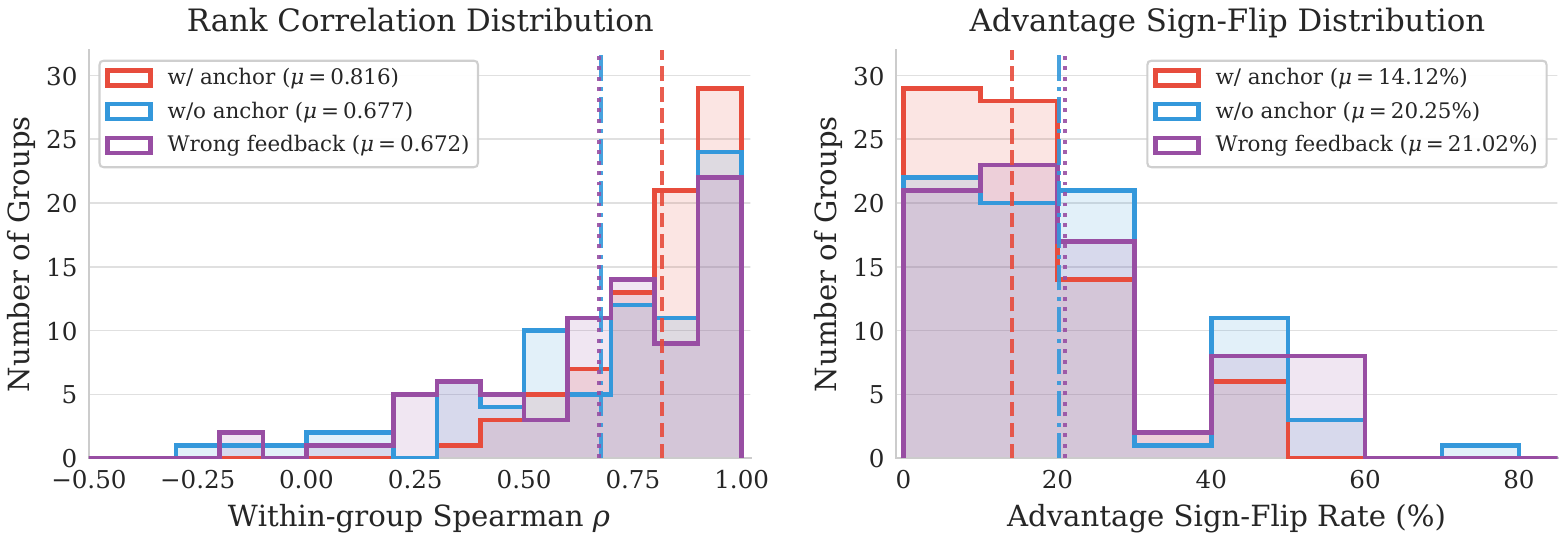}
    \caption{Within-group Spearman rank correlation (left) and per-group advantage sign-flip rate (right) across 100 randomly sampled Gemma4 task groups on DEAL-RS under \textit{w/ anchor}, \textit{w/o anchor}, and \textit{wrong feedback}.}
    \label{fig:app-anchor-3cond}
\end{figure}

\paragraph{Rank correlation and sign-flip distributions.}
Across the 100 randomly sampled task groups, Figure~\ref{fig:app-anchor-3cond} reports the within-group Spearman rank correlation (using average ranks for ties) and per-group advantage sign-flip rate.
Conditioning the rubric on the live feedback anchor achieves a mean within-group Spearman correlation of $\mu=0.816$ and a mean per-group sign-flip rate of $14.12\%$, clearly improving over both the task-only \textit{w/o anchor} variant ($\mu=0.677$, sign-flip $20.25\%$) and the \textit{wrong feedback} condition ($\mu=0.672$, sign-flip $21.02\%$).
Bootstrap confidence intervals use 4,000 paired group-bootstrap samples with seed 20260926.

\paragraph{Advantage comparison.}
For a reward vector $r$, we compute $A_i=(r_i-\bar r)/(\operatorname{Std}(r)+10^{-8})$ using the population standard deviation.
\textit{Internal-only} normalizes the seven internal self-rewards and compares them with the normalized GT rewards of those same seven trajectories.
\textit{Mixed} normalizes the live GT reward together with the seven internal self-rewards.
Its reference group contains the live GT reward and the seven internal GT rewards.
We count a sign flip when the estimated and reference advantages have strictly opposite signs.
We exclude internal trajectories with $|A_i^{\mathrm{GT}}|\le0.05$ and report the excluded fraction separately.
An estimated advantage of zero is reported separately and is not counted as a strict sign flip.

\begin{table}[htbp]
\centering
\caption{Reward-source mixing on the fixed 100-group diagnostic subset. Flip rates are over eligible internal trajectories. Exclusion rates cover all 700 internal trajectories, and live suppression covers all 100 groups.}
\label{tab:reward-grouping}
\small
\setlength{\tabcolsep}{5pt}
\begin{tabular}{llrrr}
\toprule
Self-reward & Grouping & Sign flips & GT near-zero & Live suppression \\
\midrule
w/ anchor & Internal-only & $108/681$ ($15.86\%$) & $2.71\%$ & -- \\
w/ anchor & Mixed & $127/684$ ($18.57\%$) & $2.29\%$ & $29/100$ \\
w/o anchor & Internal-only & $157/681$ ($23.05\%$) & $2.71\%$ & -- \\
w/o anchor & Mixed & $164/684$ ($23.98\%$) & $2.29\%$ & $38/100$ \\
\bottomrule
\end{tabular}
\end{table}

\paragraph{False suppression of the live trajectory.}
A live trajectory is falsely suppressed when its all-GT advantage exceeds $0.05$ but its Mixed advantage is negative.
Of the 54 groups with a clearly positive GT-reference live advantage, this occurs in 29 groups with anchor-conditioned self-rewards ($53.70\%$) and 38 groups without the anchor ($70.37\%$).
The corresponding rates over all groups are $29\%$ and $38\%$.
The Mixed-minus-Internal-only increase in internal sign flips is $2.71$ percentage points with the anchor (95\% bootstrap interval: $[0.23,5.30]$) and $0.92$ points without it ($[-0.78,2.78]$).
Thus, the latter increase is not clearly separated from zero in this sample.
Estimated zero advantages account for $12/681$ and $6/684$ eligible trajectories in the two with-anchor groups, and $11/681$ and $6/684$ without the anchor.
These results motivate keeping externally judged rewards and self-rewards out of the same normalized update group.
They do not imply that the anchor should be removed from self-reward estimation.

\subsection{Coverage Gain and Selection Loss}
\label{sec:diagnostics}
\label{app:candidate-gain-details}

We use the same seven internal candidates per group to examine the decomposition in Section~\ref{sec:theoretical-justification}.
For each $K\in\{1,\ldots,7\}$, we enumerate all $\binom{7}{K}$ subsets.
For a subset $S$, let
\[
V_{\mathrm{random}}(S)=\frac{1}{K}\sum_{k\in S}R_k,\qquad
V_{\mathrm{best}}(S)=\max_{k\in S}R_k,\qquad
V_{\mathrm{self}}(S)=\frac{1}{|J(S)|}\sum_{k\in J(S)}R_k,
\]
where $J(S)=\arg\max_{k\in S}\hat r_k$.
The last expression is the expected GT reward under uniform tie-breaking among self-reward maximizers; GT scores are never used to break the tie.
We average subsets within each group and then average the 100 groups equally.
Coverage gain is $V_{\mathrm{best}}-V_{\mathrm{random}}$, selection loss is $V_{\mathrm{best}}-V_{\mathrm{self}}$, and net candidate gain is $V_{\mathrm{self}}-V_{\mathrm{random}}$.
Net gain therefore equals coverage gain minus selection loss.
This evaluates a fixed candidate pool and does not rerun learning at different rollout budgets.

\begin{table}[!htbp]
\centering
\caption{Exact subset analysis on the fixed 100 groups. All values are GT score points on the 0--100 scale. Coverage is shared because both scoring variants use the same candidates.}
\label{tab:candidate-gains}
\small
\begin{tabular}{crrrrr}
\toprule
& & \multicolumn{2}{c}{w/ anchor} & \multicolumn{2}{c}{w/o anchor} \\
\cmidrule(lr){3-4}\cmidrule(lr){5-6}
$K$ & Coverage gain & Selection loss & Net gain & Selection loss & Net gain \\
\midrule
1 & 0.00 & 0.00 & 0.00 & 0.00 & 0.00 \\
2 & 12.17 & 2.29 & 9.88 & 4.53 & 7.64 \\
3 & 18.91 & 4.37 & 14.55 & 8.18 & 10.73 \\
4 & 23.28 & 6.15 & 17.13 & 10.94 & 12.34 \\
5 & 26.29 & 7.65 & 18.64 & 13.05 & 13.23 \\
6 & 28.46 & 8.88 & 19.58 & 14.69 & 13.77 \\
7 & 30.13 & 9.87 & 20.26 & 16.00 & 14.13 \\
\bottomrule
\end{tabular}
\end{table}

\begin{figure}[!htbp]
    \centering
    \begin{minipage}[c]{0.32\linewidth}
        \centering
        \includegraphics[width=\linewidth]{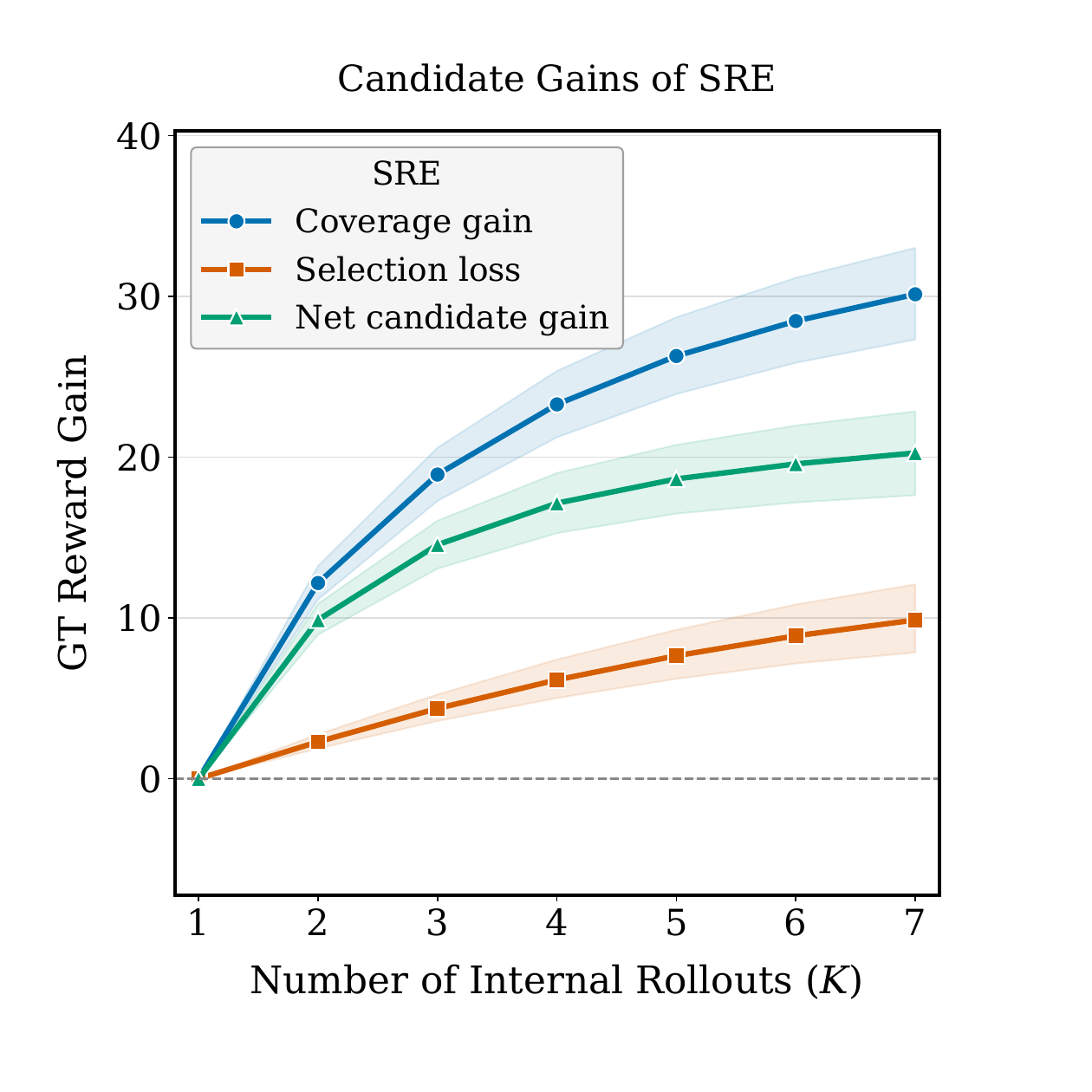}
    \end{minipage}\hfill%
    \begin{minipage}[c]{0.32\linewidth}
        \centering
        \includegraphics[width=\linewidth]{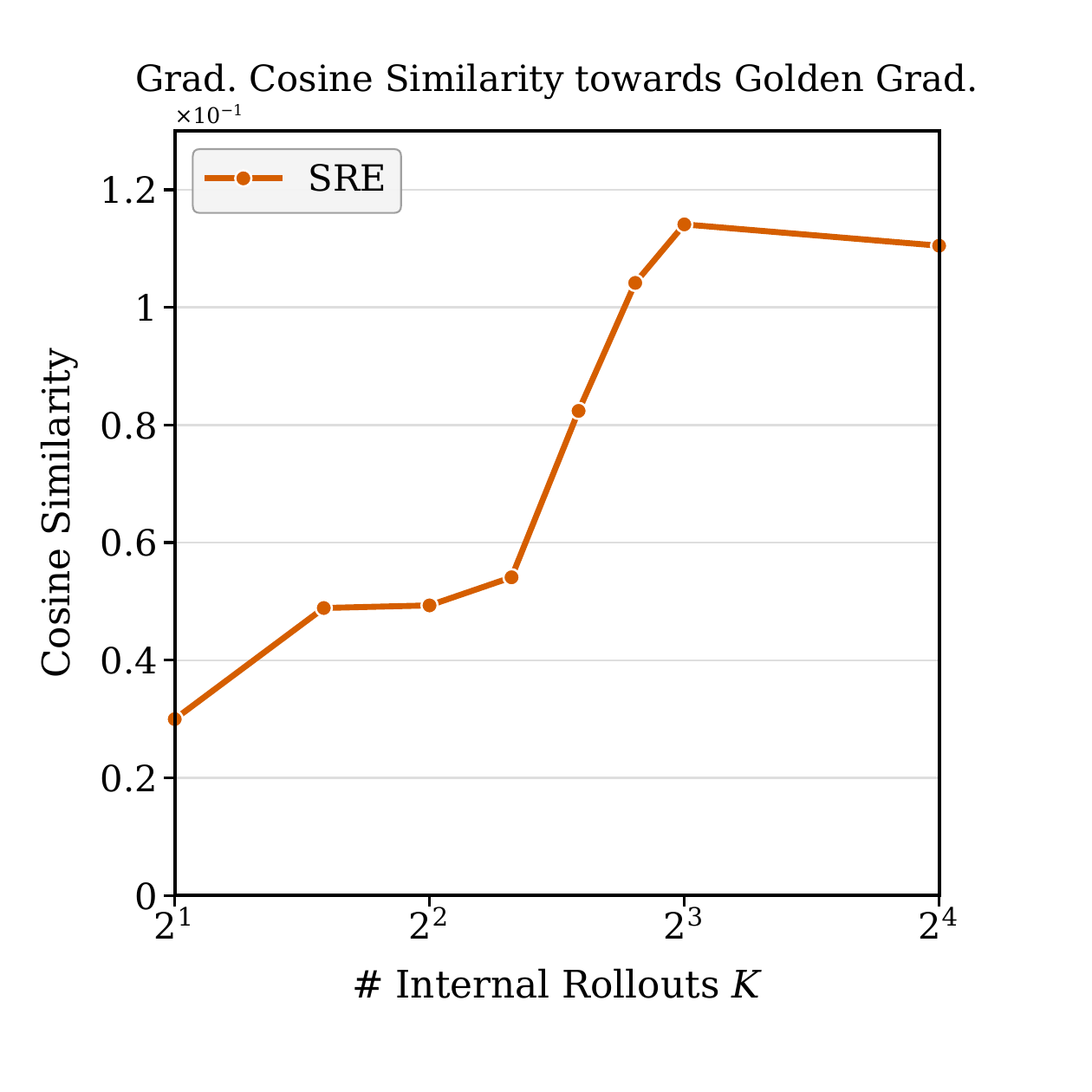}
    \end{minipage}\hfill%
    \begin{minipage}[c]{0.32\linewidth}
        \centering
        \includegraphics[width=\linewidth]{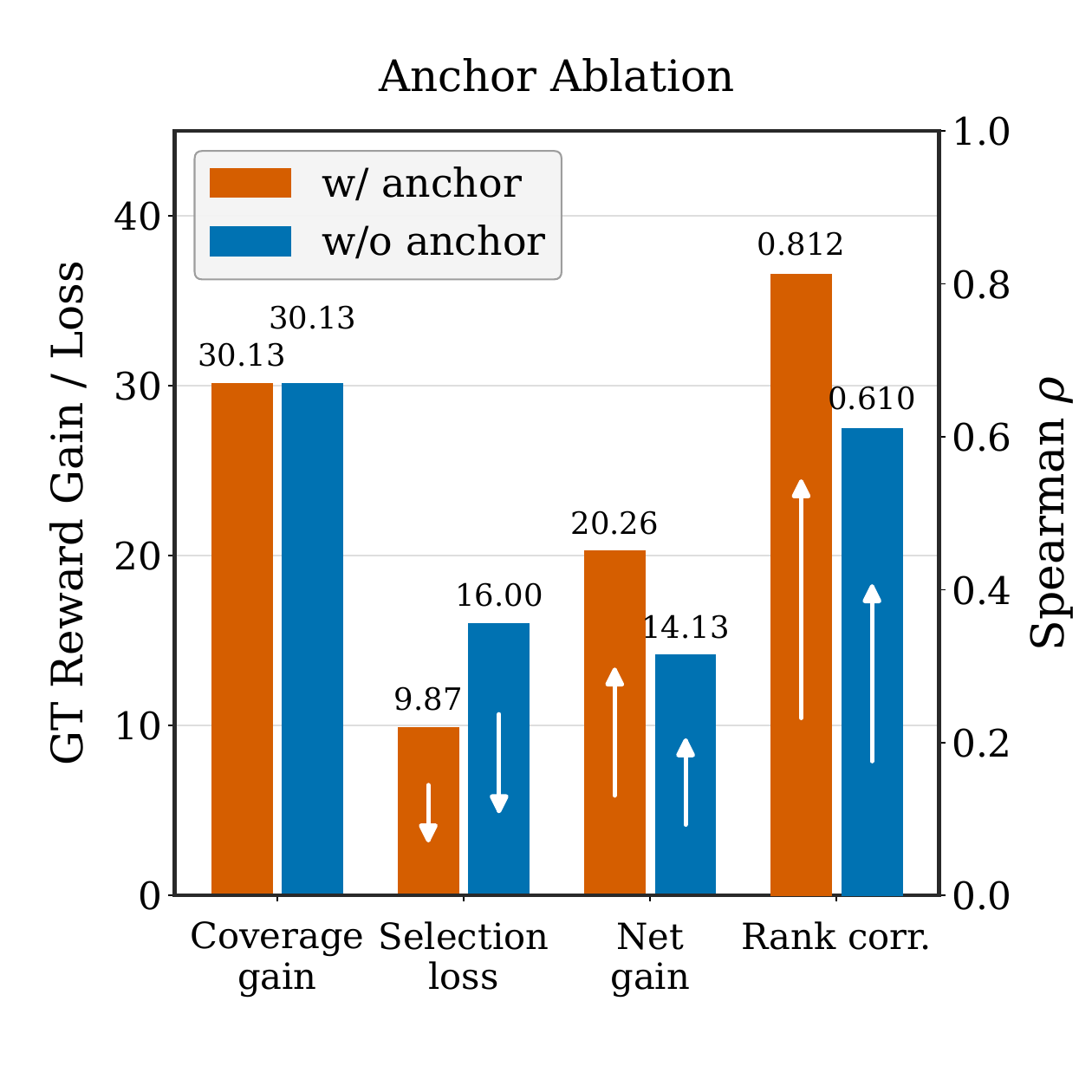}
    \end{minipage}
    \caption{Offline candidate and gradient diagnostics on the 100 randomly sampled Gemma4 task groups on DEAL-RS. Left: coverage gain, selection loss, and net candidate gain as a function of $K$. Middle: gradient cosine similarity with a 64-rollout GT-scored GRPO reference. Right: candidate selection breakdown with and without the feedback anchor at $K=7$.}
    \label{fig:app-rollout-diagnostics}
\end{figure}

Table~\ref{tab:candidate-gains} and Figure~\ref{fig:app-rollout-diagnostics} show that more candidates expose better GT-scored trajectories.
At $K=7$, $V_{\mathrm{random}}=29.93$ and $V_{\mathrm{best}}=60.06$.
Selection by self-reward achieves $V_{\mathrm{self}}=50.19$ with the anchor and $44.06$ without it, as shown in the right panel of Figure~\ref{fig:app-rollout-diagnostics}.
Scoring errors offset part of the coverage gain, but the remaining net gain is positive.
Both variants attain their highest observed net gain at $K=7$ within the examined range.
In addition, the middle panel of Figure~\ref{fig:app-rollout-diagnostics} compares the group-centered policy gradient against a 64-rollout GT-scored GRPO reference, showing that cosine similarity levels off around $K=7$--$8$ in line with Figure~\ref{fig:sre-end-to-end-ablations}(b).
In particular, $V_{\mathrm{best}}$ is an offline oracle upper bound and is not deployed performance.

\Needspace{10\baselineskip}
\subsection{General Capabilities Results}
\label{app:general-capability-details}

Table~\ref{tab:basic-ability} reports the per-benchmark breakdown of General Capabilities before and after deployment.
\begin{table}[htbp]
\centering
\caption{Basic abilities of Gemma4-E4B-IT before and after deployment training. All scores are reported on a 0--100 scale. Avg.\ is the unweighted mean over the seven benchmarks, using the harmonic mean for BBEH and prompt-level strict accuracy for IFBench. CASCADE and EvoSkill share Base under weight-only evaluation.}
\label{tab:basic-ability}
\setlength{\tabcolsep}{4pt}
\renewcommand{\arraystretch}{1.12}
\resizebox{\textwidth}{!}{%
\begin{tabular}{llrrrrrrrr}
\toprule
Training setting & Method & MMLU-Pro & BBEH & GPQA-D & AIME 2026 & MATH-500 & HumanEval+ & IFBench & Avg. \\
\midrule
Shared & Base / CASCADE / EvoSkill & 66.51 & 11.24 & 55.05 & 48.33 & 66.50 & 88.41 & 34.33 & 52.91 \\
\midrule
DEAL-RS & REINFORCE++ & 66.22 & 14.45 & 53.03 & 48.33 & 63.50 & 90.24 & 35.00 & 52.97 \\
DEAL-RS & SRE (Ours) & 67.57 & 18.65 & 58.59 & 50.00 & 66.65 & 89.63 & 35.00 & 55.16 \\
\midrule
DEAL-WS & REINFORCE++ & 66.48 & 14.07 & 54.55 & 47.50 & 66.15 & 88.41 & 33.67 & 52.98 \\
DEAL-WS & SRE (Ours) & 66.11 & 12.80 & 56.06 & 45.83 & 64.95 & 89.63 & 34.67 & 52.87 \\
\bottomrule
\end{tabular}%
}
\end{table}

\subsection{Feedback Efficiency, Imperfect Feedback, and Seed Robustness}
\label{app:additional-rebuttal-ablations}

\paragraph{Feedback efficiency across early checkpoints.}
To quantify how much supervision SRE requires before yielding tangible improvements, Figure~\ref{fig:fb-efficiency-robustness}(a) evaluates intermediate checkpoints on the DEAL-RS held-out set across the first $500$ feedback events.
Following a brief warm-up over the initial $100$ tasks ($21.30 \rightarrow 22.29 \rightarrow 21.90$), held-out performance rises steeply between $100$ and $300$ feedbacks ($28.12$ at $200$ and $32.56$ at $300$).
By $500$ feedbacks ($34.55$), SRE achieves a $+13.25$-point gain over Base and approaches the full-stream performance of REINFORCE++ trained on all $1{,}800$ tasks ($36.95$), ultimately reaching $41.22$ at the end of deployment.

\paragraph{Robustness to imperfect user feedback.}
In practical deployments, user feedback may be inaccurate, intermittent, or restricted to coarse numerical ratings without written explanations.
Figure~\ref{fig:fb-efficiency-robustness}(b) evaluates these three regimes on DEAL-RS over $20$ training steps ($320$ tasks) under clean held-out evaluation:
(1) \textit{Incorrect feedback}, where with task-level probability $p\in\{0.2, 0.5, 0.8\}$ the user simulator's feedback system prompt is replaced by a deliberately misleading instruction (prompt provided in Appendix~\ref{app:prompts}), perturbing only the textual comment $c_t$;
(2) \textit{Occasional feedback} ($50\%$ missing), where feedback is withheld on $50\%$ of randomly selected tasks and SRE falls back to its task-only \textit{w/o anchor} rubric update; and
(3) \textit{Scalar-only feedback}, where textual comments are omitted and only the scalar score $u_t$ is returned.
Across all settings, SRE improves substantially over Base ($21.30$); even under $80\%$ misleading feedback ($28.95$) or scalar-only ratings ($29.10$), it remains above the unanchored baseline ($28.23$).
Three mechanisms account for this stability: (i) the self-rubric agent inspects the original task specification $x_t$ and source files directly, preserving core task requirements in $\mathcal Q_t$ (case study provided in Appendix~\ref{app:case-studies}); (ii) whenever an anchor is available, its execution artifacts clarify concrete deliverable paths and verification targets; and (iii) the anchor is passed only to the self-rubric agent rather than the self-scorer, so any shared criterion offset $b_x$ cancels in group-centered advantages following Proposition~\ref{prop:biased-reward-exploration} and Corollary~\ref{cor:group-centered-gradient}.

\paragraph{Details of feedback-conditioned distillation (SDPO) and multi-seed robustness.}
For the SDPO baseline~\citep{hubotter2026reinforcementlearningselfdistillation} in Figure~\ref{fig:sre-end-to-end-ablations}(a), we retain the same $K=7$ offline rollouts (excluding the live anchor from gradient updates) and replace outcome-level rubric rewards with token-level teacher supervision: the frozen base model $\pi_{\btheta_0}$ conditions on $(x_t, F_t)$ as privileged context, whereas the student conditions on the task context alone.
For each assistant token $y_i$ across the seven rollouts, token-level advantages are computed as $A_i = \log \pi_{\btheta_0}(y_i \mid \text{context}, F_t) - \log \pi_{\btheta}(y_i \mid \text{context})$.
Finally, Table~\ref{tab:seed-robustness} summarizes all DEAL-RS end-to-end ablations alongside an independent run of SRE (\texttt{seed1}) with a shuffled online task arrival order.
Both runs consistently outperform REINFORCE++ on DEAL-RS ID and AARRI-Bench, showing low variance across seeds and task orders.

\begin{table}[htbp]
    \centering
    \caption{End-to-end ablations and multi-seed robustness of Gemma4-E4B-IT on DEAL-RS.}
    \label{tab:seed-robustness}
    \vspace{0.3em}
    \small
    \setlength{\tabcolsep}{7pt}
    \begin{tabular}{lrr}
        \toprule
        Method / Setting (DEAL-RS) & DEAL-RS (ID) & AARRI-Bench (OOD) \\
        \midrule
        Base & $21.30$ & $35.38$ \\
        SDPO ($K=7$, feedback-conditioned distillation) & $26.10$ & $35.37$ \\
        REINFORCE++ ($1\times$ update/step) & $36.95$ & $37.01$ \\
        REINFORCE++ (GPU-hour matched, $2\times$ updates/step) & $38.40$ & $37.43$ \\
        SRE w/o anchor ($K=7$, task-only rubric) & $37.39$ & $37.25$ \\
        SRE ($K=3$) & $33.92$ & $36.61$ \\
        SRE ($K=5$) & $36.51$ & $37.52$ \\
        \midrule
        \textbf{SRE ($K=7$, \texttt{seed0} stream)} & \textbf{41.22} & \textbf{38.60} \\
        \textbf{SRE ($K=7$, \texttt{seed1} shuffled stream)} & $40.66$ & $37.50$ \\
        \textbf{SRE ($K=7$, mean $\pm$ std across seeds)} & $40.94 \pm 0.40$ & $38.05 \pm 0.78$ \\
        \bottomrule
    \end{tabular}
\end{table}
\FloatBarrier
\subsection{Failure Analysis for Parameter-Frozen Methods}
\label{app:parameter-frozen-failures}

We examine where stored experience fails to improve execution.
Memory methods must retrieve relevant cases, while skill methods must produce, access, and follow reusable guidance.
We use Qwen3.5-9B and GLM-5.2 as the lower- and higher-performing executors in this diagnostic.
This comparison concerns these two models on the evaluated tasks, rather than general model capability.

\subsubsection{Memory-Based Methods}
\label{app:memory-failure-cases}

The memory case in Appendix~\ref{app:case-studies} shows a mismatch between project relevance and procedural relevance.
The retrieved case uses the same paper and repository as the new task, but asks for a different operation.
It therefore provides little guidance for the numerical checks required by the incoming task.
This example illustrates a retrieval limitation; it does not establish how often the limitation occurs across the stream.

\subsubsection{Skill-Based Methods}
\label{app:skill-failure-cases}

For skill methods, we separate three possible failure points: writing reusable skills, accessing the relevant skills, and executing them correctly.
Table~\ref{tab:skill-failure-ablations} lists the configurations in this diagnostic.

\begin{table}[htbp]
    \centering
    \caption{
        Skill diagnostic results on DEAL-RS. Scores use the 0--100 scale. Checkmarks indicate settings that evolve skills; crosses indicate settings without skill evolution. In A--D, the agent autonomously decides whether to read the mounted skills. E and G use no skills. F always uses a fixed set of skills generated by B, with their complete contents injected directly into the prompt; therefore, its file-read rate is not applicable.
    }
    \vspace{0.5em}
    \label{tab:skill-failure-ablations}
    \footnotesize
    \setlength{\tabcolsep}{4pt}
    \begin{tabular*}{0.9\linewidth}{@{\extracolsep{\fill}}lllcrr@{}}
        \toprule
        ID & Executor & Skiller & Evol. & Read rate & Score \\
        \midrule
        A & Qwen3.5-9B & Qwen3.5-9B & $\checkmark$ & $0.04\%$ & $57.08$ \\
        B & Qwen3.5-9B & GLM-5.2 & $\checkmark$ & $0.51\%$ & $56.75$ \\
        C & GLM-5.2 & Qwen3.5-9B & $\checkmark$ & $34.70\%$ & $87.01$ \\
        D & GLM-5.2 & GLM-5.2 & $\checkmark$ & $80.00\%$ & $87.52$ \\
        \midrule
        E & Qwen3.5-9B & None & $\times$ & $0.00\%$ & $56.04$ \\
        F & GLM-5.2 & Fixed skills & $\times$ & N/A (inline) & $86.80$ \\
        G & GLM-5.2 & None & $\times$ & $0.00\%$ & $81.13$ \\
        \bottomrule
    \end{tabular*}
\end{table}

\paragraph{Skill generation.}
In the inspected cases, the lower-performing skiller often retains details of a single task instead of writing a reusable procedure.
Its skills may omit verification steps or leave their application conditions unclear.
Replacing this skiller with GLM-5.2 produces clearer instructions under manual inspection, but does not improve the Qwen executor's task score (A vs.\ B).
Both settings remain close to the same executor without skills (E).
Better-written skills alone therefore do not explain the final task score.

\paragraph{Skill access.}
Qwen rarely reads the mounted skills in either A or B.
Skill access becomes substantially more frequent when GLM-5.2 is the executor in C and D.
The read-rate difference suggests that accessing the guidance is itself a bottleneck.
A skill cannot affect an execution through its contents if the executor does not read it.
Configuration F removes this access step by placing a fixed set of skills from B directly in the prompt.

\paragraph{Skill execution.}
Access does not guarantee that the executor follows the guidance correctly.
In the paired diagnostic, both models receive the same skill.
On the 83 tasks completed by both, GLM-5.2 scores 91.4 on average, compared with 63.0 for Qwen.
It scores higher on 65 tasks, ties on 14, and scores lower on four.
These averages are conditional on completion by both models.
The fixed-skill comparison is consistent with this observation: GLM-5.2 performs better with the injected skills in F than without skills in G.
Together, these results separate skill generation, skill access, and skill execution.
Improving the stored guidance does not remove the executor's need to collect evidence, follow task constraints, and verify its output.

\clearpage
\raggedbottom
\section{Qualitative Case Studies}
\label{app:case-studies}

We present representative traces showing how agents use prior experience and feedback. The first cases examine memory retrieval and the writing, reading, and execution of skills; the final cases illustrate how live feedback can improve self-rubric construction and help select stronger internal trajectories.
Each box reports the task, the available experience or feedback, the observed behavior, and the evaluator's diagnosis.

\subsection{Memory-Based Case}

\begin{casebox}{Same Project, Wrong Operation}
\small
\casefield{casetask}{Task}
Audit the quantitative methodology claims in \textit{Unveiling Ruby}, including survey counts, Cohen's kappa, sampling assumptions, and derived approval and relevance statistics.\par\medskip

\casefield{caseexperience}{Retrieved memory}
The retrieved successful case uses the same paper and repository, but solves a different problem: it aligns the paper's data acquisition, filtering, preprocessing, and topic extraction pipeline with the codebase.\par\medskip

\casefield{casetrace}{Execution trace}
The Executor follows the shared paper and repository context, identifies the reported kappa values, and notes several missing local inputs. However, it does not reconstruct the approval and rejection arithmetic, check the rounding of the relevance rate, or inspect the Boolean label conversion in the local kappa script.\par\medskip

\casefield{casefeedback}{Evaluator feedback}
The response receives $0.333$. The evaluator credits the reported kappa values and broad evidence gaps, but marks the required arithmetic and implementation checks as missing.\par\medskip

\par\nobreak\medskip\nobreak
\casefield{caseaccent}{Comment and takeaway}
This case illustrates why shared entities are not enough for useful memory retrieval. The reusable unit must include the operation and its checks, not just the paper or repository.
\end{casebox}

\subsection{Skill-Based Cases}

\begin{casebox}{A Narrow Skill with Little Reuse}
\small
\casefield{casetask}{Task}
The Executor must reconstruct an ablation table, compare MAE gaps across four datasets, and assess whether those gaps support the paper's claim.\par\medskip

\casefield{caseexperience}{Available skill}
The Skiller produced an \texttt{error-diagnoser} skill for a refused judge connection. It recognizes a specific network error, retries with exponential backoff, and falls back to an offline judge. These instructions do not address evidence collection, numerical reconstruction, or verification.\par\medskip

\casefield{casetrace}{Access trace}
The skill is mounted and announced in 33 executions but read in only 2. In this example, the Executor does not read it.\par\medskip

\casefield{casefeedback}{Evaluator feedback}
The response reconstructs the four MAE values correctly but omits the paper's claim, the comparison across datasets, and the caveats about statistical significance and other metrics, resulting in a score of $0.30$.\par\medskip

\par\nobreak\medskip\nobreak
\casefield{caseaccent}{Comment and takeaway}
This case links skill quality to later access: a record of one infrastructure failure offers little help with a benchmark audit. It illustrates limited reuse, but does not by itself explain every low skill-read rate.
\end{casebox}

\begin{casebox}{Relevant Guidance, Incomplete Execution}
\small
\casefield{casetask}{Task}
Audit nine TSGBench quality measures and distinguish qualitative observations within a measure family from a formal consistency test across measure families.\par\medskip

\casefield{caseexperience}{Available skill}
Both Executors receive the same task-specific benchmark-audit guidance.\par\medskip

\casefield{casefailure}{Weaker Executor}
It creates the requested CSV and lists the four shared weaknesses of the model-based measures. However, it omits the better direction of the measures, does not explain which weaknesses are addressed by ED and DTW, and conflates the analyses in Sections 6.1 and 6.4. It satisfies only $2/11$ criteria and scores $0.182$.\par\medskip

\casefield{casestrong}{Stronger Executor}
It covers all $11/11$ criteria and scores $1.0$.\par\medskip

\par\nobreak\medskip\nobreak
\casefield{caseaccent}{Comment and takeaway}
The same guidance can lead to different outcomes because execution requires capabilities beyond reading the skill. Parameter-frozen adaptation remains limited by the executor's ability to ground and verify the instructions.
\end{casebox}

\subsection{Self-Reward Cases}
\label{app:self-reward-cases}

\begin{casebox}{Single Feedback, Better Rubric}
\small
\casefield{casetask}{Task}
Audit whether a repository consistently implements the paper's three-part INR continuous function, including the residual
architecture's hidden dimension, global and group layers, and clustering-derived groups.
The output must provide structured paper-to-code comparisons with explicit source paths and alignment verdicts.\par\medskip

\casefield{caseexperience}{Live feedback}
The live response scores $0.50$.
The feedback reveals requirements that are not explicit in the task wording: the three components are a trend polynomial, a
seasonal Fourier term, and a residual adaptive group-based MLP; the first two must be connected to \texttt{block\_func.py};
and architectural claims must cite concrete local paper sources rather than an ``implied paper specification.''\par\medskip

\casefield{casefailure}{Without anchor}
The task-only rubric asks for three generic components but does not identify the trend and seasonal decomposition or require
component-level source grounding.
More importantly, it treats repository conformity as the evaluation target: the implementation must use hidden dimension 16,
one global layer, one group layer, and $K$ groups.
An answer can therefore lose credit for correctly reporting that a value is configurable or that the repository does not match
the paper.\par\medskip

\casefield{casestrong}{With anchor}
The feedback-conditioned rubric explicitly checks the trend and seasonal components against \texttt{block\_func.py}, requires
each architectural claim to be tied to a local paper source, and evaluates whether the answer compares the paper and
repository values.
It therefore permits a correct verdict to be a match, mismatch, or configuration caveat, consistent with the ground-truth
rubric.\par\medskip

\casefield{casefeedback}{Why the rubric is better}
The difference is not merely that the feedback-conditioned rubric contains more criteria; it corrects what the rubric is
intended to measure.
On the same seven rollout trajectories, Spearman correlation with ground-truth scores increases from $-0.502$ to $0.917$,
pairwise accuracy from $0.313$ to $0.875$, and top-1 agreement from $0$ to $1$.
The task-only rubric assigns its highest score to two rollouts whose ground-truth scores are only $0.25$ and $0.125$.
The feedback-conditioned rubric instead ranks the three ground-truth-best rollouts, each scoring $0.50$, highest.
In this example, the single feedback signal improves both criterion coverage and evaluation semantics.
\par\nobreak\medskip\nobreak
\casefield{caseaccent}{Comment and takeaway}
The live feedback corrects the evaluation target: an audit should report whether code matches the paper, rather than reward conformity to one assumed implementation. The anchor can improve candidate ranking by clarifying what counts as a correct answer.
\end{casebox}

\begin{casebox}{Weak Live Trace, Strong Offline Rollout}
\small
\casefield{casetask}{Task}
Audit whether LogosQ's QFT comparison against PennyLane, Qiskit, Yao.jl, and Q\# is fairly designed and whether the reported
$120$--$900\times$ and $6$--$22\times$ speedups match the local evidence.\par\medskip

\casefield{casefailure}{Live response}
The live response shifts the analysis toward VQE and Heisenberg results and omits the 1--24 qubit range, the claim of identical
circuit configurations, the backend switch at 11 qubits, the absence of per-qubit baseline timings, and the relationship
between the speedup ranges and the reported algorithmic complexities.
It receives a ground-truth score of $0.30$.\par\medskip

\casefield{caseexperience}{Feedback-derived evaluation signal}
The live feedback identifies these omissions.
The resulting self-rubric checks the reported execution times and speedup ranges, the switch from an FFT/state-vector
implementation at 1--10 qubits to an MPS backend at 11--24 qubits, the associated complexity claims, and the traceability of
conclusions to local paper evidence.
The feedback thus converts the failure of one trajectory into criteria that can be applied to new trajectories.\par\medskip

\casefield{casestrong}{Best additional rollout}
One of seven additional rollouts produces a substantially stronger audit.
It identifies the 11-qubit backend switch as a comparability confound, distinguishes aggregate speedup ranges from the missing
per-qubit and per-baseline timing data, connects the $120$--$900\times$ claim to the MPS complexity argument, and preserves
the caveat that the Q\# claim is qualitative rather than quantitative.
Its only ground-truth omission is failing to use the required methodology source for the baseline list and the 1--24 qubit
range.
It receives a ground-truth score of $0.90$.\par\medskip

\casefield{casefeedback}{Self-reward selection}
For this rollout, the three self-reward evaluations are $0.50$, $0.75$, and $0.75$, giving a median score of $0.75$.
This is the highest aggregate self-reward among the seven additional rollouts.
The selected rollout is also the one with the highest ground-truth score.
Across the full rollout pool, self-reward achieves Spearman correlation $0.875$, top-1 agreement $1$, and zero top-1 regret.
\par\medskip

\casefield{casetrace}{Why the rollout is better}
The improvement comes from correcting the scope of the audit and completing its evidence chain.
Unlike the live trajectory, the selected rollout remains focused on QFT, identifies the central backend-switching confound and connects the numerical claims, missing evidence, and theoretical explanation into a coherent fairness judgment.
Offline expansion supplies the stronger trajectory, while self-reward supplies the selection mechanism that recovers it from
the rollout pool.
\par\nobreak\medskip\nobreak
\casefield{caseaccent}{Comment and takeaway}
A weak live outcome can still supply useful criteria for judging new trajectories. This example illustrates both coverage gain and successful self-reward selection. The internal score of 0.90 is an offline diagnostic, not a replacement for the delivered live score of 0.30.
\end{casebox}

\begin{casebox}{Rubric Grounding Under Misleading User Feedback}
\small
\casefield{casetask}{Task}
In a paired diagnostic on DEAL-RS (\texttt{task\_0710}), the agent must reconcile a paper's relation-extraction components against local repository evidence, deliver \texttt{answer.csv} comparing TKRE vs.\ TYP Marker and TKRE-GenPT vs.\ GenPT with a column checking the paper's claimed $7.8\%$ or $5.0\%$ deltas, and write supporting caveats to \texttt{caveats.md}.\par\medskip

\casefield{casefailure}{Misleading feedback}
A deliberately misleading feedback paragraph (generated with the wrong-feedback prompt in Appendix~\ref{app:prompts}) tells the candidate to rewrite the output as a detailed Markdown report \emph{instead of} a CSV, substitute a fabricated $12.5\%$ figure for $7.8\%$, and report a fabricated GenPT average of $65.0$.\par\medskip

\casefield{casestrong}{Generated self-rubric}
Because the self-rubric agent reads both \texttt{TASK.md} and the local source files alongside \texttt{ANCHOR.md}, the resulting 5-item rubric still requires both \texttt{answer.csv} and \texttt{caveats.md} (\texttt{c1}), the required comparison and summary rows (\texttt{c2}), and explicit verification of the paper's claimed $7.8\%$ or $5.0\%$ deltas (\texttt{c3}), without adopting the fabricated format change or $12.5\%$ target.\par\medskip

\par\nobreak\medskip\nobreak
\casefield{caseaccent}{Boundary and counterexample}
Task-contract grounding reduces error propagation locally rather than guaranteeing universal immunity. In a paired counterexample (reconstructing SWIFT Table~7 loss rankings, where \texttt{TASK.md} defines Rank~1 as the highest average), when misleading feedback explicitly instructs reversing the order so the lowest average is Rank~1, the generated rubric adopts the inverted ranking in one criterion while keeping the remaining checks intact. Together with the group-centered offset cancellation in Proposition~\ref{prop:biased-reward-exploration}, these two mechanisms explain why SRE degrades gracefully rather than collapsing under noisy feedback.
\end{casebox}
\clearpage
\section{Core Prompt Templates and Output Schemas}
\label{app:prompts}

This section summarizes the core task-specific prompt templates, output schemas, and imperfect-feedback instructions used in SRE. Runtime workspace paths, tool schemas, and multi-turn tool outputs are populated dynamically by the harness.

\subsection{Self-Rubric Construction Prompts and Schema}
\label{app:prompts-rubric}

For each task, the self-rubric workspace mounts \texttt{TASK.md} (the original user request), \texttt{ANCHOR.md} (packaging the live final reply, textual user feedback $c_t$, and numeric score $u_t$), \texttt{WORKSPACE\_MANIFEST.json}, \texttt{RUBRIC\_SCHEMA.json}, and isolated views of the public task inputs and anchor deliverables (excluding any ground-truth rubric or hidden reference answer). Below we show the core prompt used on DEAL-RS; the DEAL-WS prompt follows the same structure and output contract, adding workspace path mappings (\texttt{input/...}) and the file-inspector CLI (\texttt{python tools/inspect\_workspace\_file.py <path>}) for Office/PDF files.

\begin{promptbox}{Self-Rubric Agent Prompt (DEAL-RS)}
\scriptsize\ttfamily\raggedright
\# Role\\
You are a reward agent. Design an objective rubric for grading another agent's completed ResearchClaw task. Keep the rubric truth-grounded: prefer criteria that can be decided from task-local facts or observable candidate artifacts, and avoid subjective criteria based on style, persuasiveness, or overall impression.\par\vspace{1pt}
\# Inputs\\
- TASK.md: the original user request and output contract.\\
- ANCHOR.md: one sampled candidate answer plus the user's feedback on it.\\
- WORKSPACE\_MANIFEST.json: the candidate deliverables and research sources available in this workspace.\\
- input\_files/: the original public task inputs and research sources, preserving their paths below this directory. Candidate deliverables remain at their requested output paths outside input\_files/.\\
- RUBRIC\_SCHEMA.json: the exact output format.\par\vspace{1pt}
\# Required procedure\\
1. Read TASK.md, ANCHOR.md, WORKSPACE\_MANIFEST.json, and RUBRIC\_SCHEMA.json.\\
2. Understand what the task asks the candidate to deliver and inspect the workspace files relevant to those requirements, applying the input-path mapping (P -> input\_files/P for original inputs; candidate output paths unchanged).\\
3. Use the task, anchor, feedback, candidate deliverables, and available research sources together to identify important success conditions and failure modes. The feedback is useful evidence but should not be the rubric's entire coverage.\\
4. Write 4--12 non-overlapping atomic criteria. Each criterion must describe one concrete, objectively decidable property of the candidate's work. Split requirements that can fail independently.\\
5. For every criterion, specify: criterion (observable property), expected (clear pass condition with concrete values, formulas, accepted sets, or source evidence), target\_paths (workspace deliverable files to inspect), reason (material task/user requirement covered), and weight (1.0).\\
6. Cover the important factual outputs, requested analysis, and concrete caveats without rewarding style, confidence, effort, polish, or overall impression.\\
7. Do not grade the anchor, feedback, task wording, scorer, or a hidden/reference answer as if it were the candidate's work.\\
8. Use at most 20 tool calls. Do not call goal-management tools.\\
9. Write rubric.json (schema\_version: "self\_reward\_rubric\_v3", criteria: [c1..cN], limitations: []), validate with `python -m json.tool rubric.json >/dev/null`, and do not create or modify any other file.
\end{promptbox}

\subsection{Self-Scoring Prompts and Aggregation Rules}
\label{app:prompts-scoring}

The self-scoring agent evaluates each offline rollout against the compiled \texttt{rubric.json} without receiving \texttt{ANCHOR.md} or raw user feedback $F_t$ (though feedback-informed requirements may enter indirectly through \texttt{rubric.json}). Because research and workspace tasks differ in artifact structure, the two domains use tailored native scoring harnesses:
\begin{itemize}[leftmargin=1.4em, itemsep=1pt, topsep=2pt]
    \item \textbf{DEAL-RS (two-stage hybrid judge):} Pass~1 runs a tool-free triage over the task, candidate outputs, and compiled binary ($0/1$) criteria, directly scoring criteria verifiable from the prompt and flagging remaining source-backed claims in \texttt{needs\_evidence}. Pass~2 runs a bounded targeted verifier (up to $30$ tool calls, local workspace only) to inspect \texttt{input\_files/} for the flagged criteria and write \texttt{\_judge/judge\_result.json}. Scores are averaged equally (weight $1.0$).
    \item \textbf{DEAL-WS (artifact-grounded native judge):} The scorer inspects only the candidate's final workspace artifacts and original sources in \texttt{workspace/input/} (via file tools and \texttt{inspect\_workspace\_file.py} with context-bounded previews), without reading the solver's execution trajectory. It scores each rubric item in $[0,1]$ under keys \texttt{"1"}, \texttt{"2"}, \dots\ in \texttt{judge\_result.json}, and runtime normalization computes their unweighted mean.
\end{itemize}

\begin{promptbox}{Self-Scoring Core Instructions (DEAL-RS Hybrid \& DEAL-WS Native)}
\scriptsize\ttfamily\raggedright
{[DEAL-RS Pass 1: Tool-Free Triage]} You are the first-pass ResearchClaw judge (no tools). Assign direct 0/1 scores where the prompt contents suffice, and request targeted local evidence (needs\_evidence + suggested\_paths) only where the solver made a concrete source-backed claim that must be verified against bundled input\_files/. Respect each criterion's named output path; score 0 if required content is clearly omitted. Return JSON with keys: scores, needs\_evidence, evidence\_requests, notes.\par\vspace{1pt}
{[DEAL-RS Pass 2: Targeted Verifier]} Verify only the criteria selected in Pass 1 using local reads/grep on input\_files/ (at most 30 tool calls; no network). Compare the solver's actual values and claims against local sources for omissions, label errors, or arithmetic mismatches, and write \_judge/judge\_result.json with keys: scores, notes.\par\vspace{1pt}
{[DEAL-WS Native Scorer]} Evaluate only the final artifacts and their grounding in workspace/input/ against task.md and rubric.json. Do not use the internet, inspect the evaluated agent's trajectory, or introduce hidden criteria. Inspect existing outputs (output:N) first via inspect\_workspace\_file.py with bounded previews, consult original sources (source:N) only when needed for factual verification, score every rubric item in [0.0, 1.0], and write judge\_result.json.
\end{promptbox}

\paragraph{Validity gating and three-repeat aggregation.}
Each offline rollout is scored in $3$ independent self-scoring sessions. Valid scalar self-rewards (including valid $0.0$ scores) are aggregated by taking the median when all $3$ sessions succeed, the mean when $2$ succeed, or the single valid score when $1$ succeeds. If all $3$ scoring sessions fail (e.g., timeout, missing/incomplete result JSON, or invalid schema), that trajectory is marked invalid and masked out from GRPO advantage normalization and policy gradient updates; if anchor execution or self-rubric generation fails, the group is masked.

\subsection{Additional Prompt Instructions for Imperfect-Feedback Ablations}
\label{app:prompts-ablations}

For the imperfect-feedback ablations in Figure~\ref{fig:fb-efficiency-robustness}(b) and Appendix~\ref{app:additional-rebuttal-ablations}, held-out evaluation uses clean ground-truth rubrics, while online deployment modifies only the feedback channel:
\begin{itemize}[leftmargin=1.4em, itemsep=1pt, topsep=2pt]
    \item \textbf{Incorrect feedback ($p\in\{0.2, 0.5, 0.8\}$):} For each online task, we draw a deterministic hash $u = \mathrm{SHA256}(\texttt{seed} \mathbin{\|} x_t)_{0:8} / 2^{64}$ with fixed seed; when $u < p$, we replace the simulator's feedback prompt with the misleading instruction below, perturbing only $c_t$.
    \item \textbf{Occasional feedback ($50\%$ missing) \& Scalar-only feedback:} Under occasional feedback, $(u_t, c_t)$ is withheld on $50\%$ of tasks and SRE falls back to its task-only \textit{w/o anchor} rubric prompt (which omits \texttt{ANCHOR.md} and constructs criteria solely from \texttt{TASK.md} and public inputs). Under scalar-only feedback, the textual comment $c_t$ in \texttt{ANCHOR.md} is left empty while retaining the live deliverables and scalar score $u_t$.
\end{itemize}

\begin{promptbox}{Deliberately Incorrect User-Feedback System Prompt}
\scriptsize\ttfamily\raggedright
You are generating deliberately incorrect user feedback for a controlled research experiment. Use the original request, evidence packet, and candidate-specific notes to understand the task and what the candidate actually did.\par\vspace{1pt}
Write ONE natural paragraph of 3--5 sentences and approximately 60--100 English words. Address the candidate directly in a concise, confident user voice.\par\vspace{1pt}
Give substantively wrong or misleading feedback, with concrete requests to revise the work in an incorrect direction. The errors must conflict with the supplied task or evidence; a mere change in tone or preference is not enough. You may introduce multiple errors, and there is no restriction on their type or scope. Errors may concern arithmetic, ranking or metric direction, numeric or factual claims, file or table existence, evidence interpretation, generalization, theoretical applicability, or a reversal of the overall assessment. Broad claims such as applicability to all settings are also allowed. These are examples, not a checklist: choose freely rather than forcing every error type into each response.\par\vspace{1pt}
Keep the feedback connected to the actual task and deliverable. It may mix correct and incorrect statements, but do not dilute it into mostly correct feedback. Do not explain or reveal that the feedback is intentionally wrong. Avoid lists, long recaps, and unnecessary detail.\par\vspace{1pt}
Return exactly \{"feedback": "..."\}, with no other keys or text. Only the feedback text is being perturbed; do not output or modify any numeric reward field.
\end{promptbox}
\end{document}